\documentclass[runningheads]{llncs}
\usepackage{graphicx}
\usepackage{comment}
\usepackage{amsmath,amssymb} %
\usepackage{color}

\usepackage{multirow}
\usepackage{diagbox}
\usepackage{xspace}
\usepackage{subfigure}
\usepackage{placeins}

\makeatletter
\DeclareRobustCommand\onedot{\futurelet\@let@token\@onedot}
\def\@onedot{\ifx\@let@token.\else.\null\fi\xspace}

\def\eg{\emph{e.g}\onedot} 
\def\ie{\emph{i.e}\onedot}

\def\etal{\emph{et al}\onedot}
\makeatother

\begin{document}
\pagestyle{headings}
\mainmatter

\title{Flexible Example-based Image Enhancement \\
with Task Adaptive Global Feature Self-Guided Network}

\titlerunning{Example-based Image Enhancement with MT-GSGN}

\author{Dario Kneubuehler \and
Shuhang Gu \and
Luc Van Gool \and
Radu Timofte}

\authorrunning{D. Kneubuehler et al.}
\institute{Computer Vision Lab, ETH Zurich, Switzerland}

\maketitle

\begin{abstract}
We propose the first practical multitask image enhancement network, that is able to learn one-to-many and many-to-one image mappings. We show that our model outperforms the current state of the art in learning a single enhancement mapping, while having significantly fewer parameters than its competitors. Furthermore, the model achieves even higher performance on learning multiple mappings simultaneously, by taking advantage of shared representations. Our network is based on the recently proposed SGN architecture, with modifications targeted at incorporating global features and style adaption. Finally, we present an unpaired learning method for multitask image enhancement, that is based on generative adversarial networks (GANs).

\end{abstract}

\section{Introduction}
\label{sec:introduction}
Digital images are omnipresent in today's society, with a wide scope of applications ranging from posting snapshots taken with smartphones on social media, to high profile fashion shoots and photojournalism.
Current image enhancement software provides tools to locally and globally adjust images to one's liking, yet the use of such tools requires a considerable amount of time and the results highly depend on the user's skills. Automating this work through the use of algorithms remains a challenging task to this day. 
In this paper, we study the problem of automatic image enhancement through deep neural networks (DNN).
Image enhancement comprises a wide set of different image processing tasks ranging from image denoising, super resolution to illumination adjustment.
We focus on the task of example-based image enhancement, which aims to enhance a group of low-quality images to the quality of another group of high quality images.
Early image enhancement works~\cite{aubry2014fast,farbman2008edge,wang2013naturalness} mainly focused on contrast enhancement and illumination estimation using algorithmic models and traditional machine learning methods.
Recently, the success of deep neural networks (DNN) on other computer vision tasks triggered the study of DNN-based image enhancement approaches.
Compared with traditional heuristic or prior based approaches, DNN methods that directly learn a mapping function from the low quality image to high quality image, have achieved promising enhancement performance.
Despite their impressive performance, current DNN-based methods are limited by the following three issues.

\textbf{Network architecture}. Compared with classical image restoration tasks such as image denoising and super-resolution, the image illumination enhancement tasks requires global adjustment of the input and thus a very large receptive field is inevitable. 
Although several network architectures have been proposed for this task, better architectures are necessary to achieve a better trade-off between performance and efficiency.

\textbf{Limited flexibility}. The working scenarios of enhancement algorithms are very complex, which requires the enhancement network to be highly flexible.
On the one hand, low quality images produced by different devices, or by the same device in different environmental conditions have distinct characteristics. The poor generalization of existing DNN-based algorithms requires us to train different networks for different types of degradation, which is highly impractical.
On the other hand, image enhancement is a highly subjective task, for the same input image, different people might favor different enhancement results. 
Adapting to the preference of different customers is very important.

\textbf{Training data}. Enhancement algorithms need to adjust images globally and locally. In order to learn a global illumination mapping a large number of training samples is required to provide image-level supervision. 
However, the acquisition of paired data for image enhancement is laborious and costly.
In ~\cite{fivek}, Bychkovsky~\etal\ provide a dataset with 5000 raw input images and corresponding professionally retouched versions, that is well suited for use with supervised machine learning methods. However, 5000 pairs of images are far from enough to train a good enhancement network.
In ~\cite{ignatov2017dslr}, Ignatov~\etal\  collected weakly paired data from different devices, \eg, cell phones and DSLR camera. However, the authors only provide roughly aligned patches (100 $\times$ 100px) which significantly limits the valid receptive field of an enhancement network in the training phase.

In this paper, we propose a novel network architecture to address all the issues above.
To efficiently exploit large scale contextual information, we modify the recently proposed self-guided network~\cite{SGN_ICCV} by adding multiple modification targeted at incorporating global features.
Specifically, we follow the top-down strategy of SGN and incorporate large scale contextual information at an early stage to guide the following processing steps, by adding a global feature branch on top of it.
Besides incorporating global information via the global feature branch, we constantly incorporate image scale information using instance normalization (IN) layers on all levels of the network.
Concretely, for handling different enhancement tasks with a compact model, we switch the instance normalization layers with adaptive instance normalization Layers (AdaIN) and add a small auxiliary network that transforms the latent input.
Using this scheme we train the network for different tasks and only allow the AdaIN layers to be task-specific.
Another advantage of such a multi-task learning strategy lies in a data augmentation perspective.
As the network is able to take advantage of shared feature representations and leverage training data from multiple tasks, our method achieves better results on both the many-to-one mapping and one-to-many mapping cases.
In summary, our main contributions are:
\begin{itemize}
  \item We propose a novel SGN~\cite{SGN_ICCV} based deep neural network architecture that outperforms its competitors by a large margin on the task of supervised image to image mapping, while requiring considerably fewer parameters.

  \item We show that by using multitask learning to learn multiple mappings simultaneously, we can take advantage of shared representations which yields even higher performance compared to learning one-to-one mappings separately. 
 
  \item For the supervised and unsupervised settings, we conduct experiments for both, one-to-many and many-to-one mappings. 
  The experimental results validate the effectiveness of multi-task learning for image enhancement.
\end{itemize}

\section{Related Work}
\label{sec:related_work}

Image enhancement is a classical computer vision problem.
While initial research in the field was based on algorithms that rely on heuristic rules, such as histogram equalization and retinex-based methods, recent research has shifted to learning based methods.
Specially, as a powerful tool for image to image mapping, deep neural networks (DNNs) have achieved great success in image enhancement.
Previous works have investigated different aspects of DNN-based image enhancement.
One category of studies aims to investigate better network for capture the mapping function between input and target image.
Yan~\etal~\cite{Yan2014} trained a multilayer feed forward neural network to capture the mapping function between two groups of images.
Lore~\etal~\cite{LoreAS15} used an autoencoder based network to tackle low-light image enhancement.
Gharbi~\etal~\cite{gharbi2017deep} presented a bilateral learning approach that is optimized for real time performance on smartphones.
Chen~\etal~\cite{DPE} augmented the U-Net approach~\cite{U-Net} with a global branch to better capture the global information.
Wang~\etal~\cite{Wang_2019_CVPR} presented a supervised learning method for improving underexposed photos that relies on an intermediate illumination mapping.
Another category of research attempts to push image enhancement toward real application scenarios.
To improve the quality of cell phone images,
Ignatov~\etal~\cite{ignatov2017dslr} collected images with different devices and roughly aligned image patches for training enhancement network.
However, as accurate image registration is a challenging problem, Ignatov~\etal~\cite{ignatov2017dslr} only provide image patches for training, which greatly limited the capacity of network to leverage global information.
To enhance images for real applications, unsupervised learning approaches have been proposed.
Deng~\etal~\cite{DengLT17} used adversarial learning for aesthetic-driven image enhancement with weak supervision.
Chen~\etal~\cite{DPE} proposed a deep photo enhancer which 
relies on a two-way generative adversarial network (GAN) architecture.

While some of the recently proposed models show impressive results for supervised and weakly supervised image enhancement, none of them is able to learn more than a single deterministic mapping function. However, users of enhancement software are known for having different aesthetic preferences, that cannot be satisfied with a single mapping function. This limitation is also non-ideal from the perspective of neural network training, since a lot more data would be available if the models were able to learn multiple tasks simultaneously. The next logical step is to introduce multi-task learning into image enhancement and provide the user with high level controls that can satisfy multiple stylistic preferences.

\begin{figure*}
\begin{center}
    \includegraphics[width=0.8\linewidth]{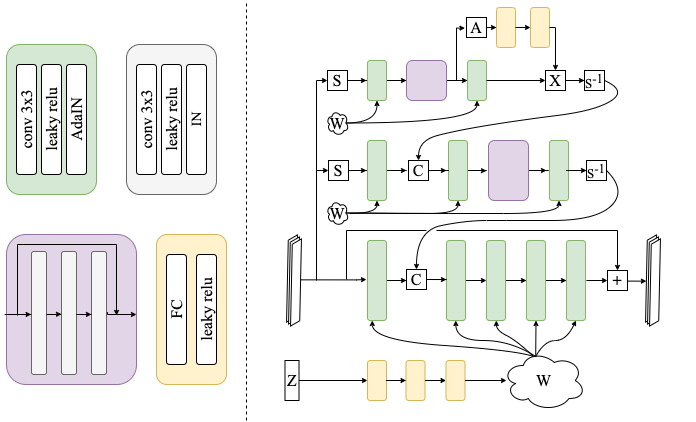}
\end{center}
    \caption{Architecture of MT-GSGN network. The latent vector $z$ is mapped to a intermediate latent space $w$, which is used as input to the adaptive instance normalization layers (AdaIn). For GSGN no mapping network is used and AdaIn layers are replaced with instance normalization (IN) layers. $S$ stands for the shuffling operation, $S^{-1}$ for the inverse shuffling operation, and $C$ for concatenation. $A$ denotes global average pooling.  Residual blocks, convolutional blocks and fully connected blocks, have purple, green and orange color coding respectively.
    }
    \label{fig:modified_sgn}
\end{figure*}

\section{Proposed Method}
\label{sec:proposed_methods}

In this section, we introduce our Global feature Self-Guided Network (GSGN) model and how it can be extend to the multi-tasks version, \ie MT-GSGN.
First, we briefly review the network architecture of the original SGN model.
Then, we introduce modifications targeted at solving the illumination enhancement problem.
Lastly, we present how the task-specific global feature branch can be utilized, and propose the flexible MT-GSGN method.

\subsection{Brief introduction to SGN~\cite{SGN_ICCV}}
\label{ssc:SGN}

Gu~\etal~\cite{SGN_ICCV} proposed the Self-Guided Network (SGN) to more efficiently incorporate large scale contextual information for image denoising. 
In order to have an overview of the image content from large receptive field, the SGN method adopts a top-down guidance strategy.
Specifically, shuffling operations are adopted to generate multi-resolution inputs, and SGN firstly processes the top-branch and gradually propagates the features extracted at low spatial resolution to higher resolution sub-networks.
With the effective self-guided mechanism, the SGN~\cite{SGN_ICCV} has achieved state-of-the-art denoising performance.

\subsection{Global Feature Self-Guided Network (GSGN)}
\label{ssc:GSGN}

As SGN has been shown to be an effective network architecture for incorporating large scale contextual information, we adopt it to solve the image illumination enhancement task, for which a large receptive field is even more essential.
Although the receptive field of SGN is much larger than the other denoising networks, it is still insufficient for the illumination enhancement task.
Furthermore, SGN has a considerable number of parameters, training it with the limited number of samples in the enhancement datasets might lead to over-fitting during training or unused networks capacity.
In order to adapt the network to the enhancement task, we introduce the Global feature Self-Guided Network (GSGN).
The architecture of the proposed GSGN is shown in Fig.~\ref{fig:modified_sgn}.
GSGN differs from SGN in the following aspects.

\textbf{Global Feature Incorporation.}
We incorporate global features by using global average pooling in the top most branch of GSGN, followed by two fully connected layers. These global features are then multiplied with the output feature maps on the same branch. In contrast to the global feature scheme used in \cite{DPE}, our approach works for arbitrary size input images.

\textbf{Less Parameters.} In order to reduce the number of parameters, we reduce the number of levels from 3 to 2 and reduce the number of channels in the higher level sub-networks by a factor of two. At the same time we increase the number of feature maps in the base level sub-network by a factor of two. 

\textbf{Constantly Incorporating Global Information.}
In spite of incorporating the global information in an early stage of the network, we employ instance normalization (IN) layers after the activation functions of convolutional layers.
While instance normalization is mostly used as a means for stabilizing training, we use it to learn global features in all levels of the network. Each instance normalization layer has two parameters for each channel. These parameters operate on the whole feature map of a single image and therefore act as global features.

With the above modifications we obtain a relatively lightweight GSGN model with only 339k parameters.
In section~\ref{ssc:single_task}, we provide an ablation study to validate our design choices.

\subsection{Flexible Image Enhancement with Task Adaptive GSGN}
\label{ssc:task_adaptive_GSGN}

To enable task adaptive learning, GSGN is augmented with an additional mapping network that consists of multiple fully connected layers with leaky relu activation functions, similar to the technique used in~\cite{stylegan2018}. The number of fully connected layers was empirically set to 3. The mapping networks takes a latent vector $z$ as input, which encodes the desired style to be learned, and maps it to a intermediate latent space $w$. This transformed latent vector is then used as the input to adaptive instance normalization~\cite{ADAIN} layers that are inserted after the convolution layers. Furthermore, an additional fully connected layer is used before each adaptive instance normalization layer to match the dimension of $w$ to the dimension of the relevant feature map.%

\subsection{Unsupervised Flexible Enhancement with Task Adaptive GSGN}
\label{ssc:task_adaptive_GSGN_unsupervised}

\begin{figure*}[h!]
\begin{center}
    \includegraphics[width=0.8\linewidth]{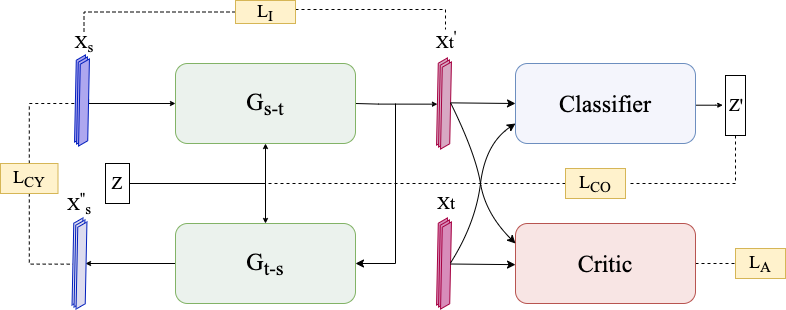}
\end{center}
    \caption{Two way GAN training setup used for learning one-to-many mappings with unpaired data. Note that for simplicity only the cycle loss from $X_s$ to $X_s^{''}$ is displayed, while the second cycle from $X_t$ to $X_t^{''}$ is omitted}
    \label{fig:traning_setup_12m}
\end{figure*}

To show the suitability of our Task Adaptive GSGN for unsupervised learning, we use the popular CycleGAN architecture~\cite{Cyclegan}, and add an addition network that provides a conditional loss based on the task.
A illustration of our unpaired training setup is shown in figure~\ref{fig:traning_setup_12m}.
In general, the setup can be described as a two player game where player one, called the generator $G$ tries to conditionally produce fake samples $X_t'$ given a prior sample $X_s$. Player two, called the critic $D$ tries to evaluate how close the fake samples $X_t'$ are to the real samples $X_t$.
In our case $X_s$ are image samples from the source distribution, while $X_t$ are high quality samples from the target distribution and $X_t'$ is the estimated enhanced version of the input image.

In addition to the generator $G_{s \rightarrow t}$ that maps samples from the source domain to the target domain, a CycleGAN architecture uses an additional generator $G_{t \rightarrow s}$ that maps samples from the target domain back to the source domain.
Using these mappings, we define two cyclic mappings between the domains: 1) $X_s \rightarrow  X_t' \rightarrow X_s''$ and
2) $X_t \rightarrow  X_s' \rightarrow X_t''$.
Since both $G_{s \rightarrow t}$ and $G_{t \rightarrow s}$ are optimized using the same loss function, we only use the term $G$ in reference for both networks.

\textbf{Adversarial Loss}
We use a version of the WGAN-GP loss ~\cite{WGAN_GP}, where a critic is used instead of a discriminator. In contrast to loss functions based on the original GAN formulation~\cite{goodfellow2014generative}, the output of the network is not feed through a sigmoid function but used directly to approximate the wasserstein distance between two probability distributions. In order for this to work the network needs to be constrained to a 1-Lipschitz function. This is achieved by putting a penalty on the gradients of the critic during training.
The gradient penalty $\lambda$ is computed by evaluating the gradients of linearly interpolated samples $\hat{y}$ between $X_t$ and $X_t'$:
\begin{equation}
    \lambda = max(0, ||\nabla \mathbb{D}(\hat{y}) ||_2 -1).
\end{equation}
While, the loss functions for the critic $D$ amounts to:
\begin{equation}
    L_{D} = ( D(X_t) - D(X_t') ) \lambda w,
\end{equation}
where $w$ is a hyper-parameter used to control the amplitude of the gradient penalty $\lambda$.
The adversarial loss for the generator $G$ can be computed as:
\begin{equation}
    L_{A} = D(X_t').
\end{equation}

\textbf{Cycle Loss}
To ensure cycle consistency of the mappings between distributions, we impose a cycle loss between $X_s$ and  $X_s''$ as well as between $X_t$ and $X_t''$. Defined as:
\begin{equation}
    L_{CY} = MSE(X_s, X_s'') + MSE(X_t, X_t'')
\end{equation}

\textbf{Illumination invariant identity loss}
The use of identity losses between the source and target images is common in CycleGAN setups, to ensure that the content of the processed image is similar to the input image.
However, since the change in illumination between input and output is large for the image enhancement task, traditional MSE and MAE pixel based losses result in high identity losses for target mapping.
Because illumination enhancement results in a shift of the mean pixel values, a simple measure to mitigate this problem is to substract the mean over the pixel values of each image before computing the identity loss.
Using the this definition, the identity loss can be stated as follows:
\begin{equation}
    L_I = MSE\left(X_s - \mu(X_s) , X_t' - \mu(X_t)\right) +  MSE\left(X_t - \mu(X_t), X_s' - \mu(X_s')\right),
\end{equation}
where $\mu$ is the arithmetic mean function, applied over all the pixel values of an image.

\textbf{Conditional Loss}
For the one-to-many case, we use an additional network $C$ that acts as a classifier on the images produced and provides a loss based on reconstructing the latent vector $z$ from the generated image. Therefore we define the conditional loss:
\begin{equation}
L_{CO} =  - \big(z \log(C(X_t')) + (1 - z) \log (1-C(X_t'))\big).
\end{equation}
The network $C$ is optimized using the same loss function but computed over $X_t$ instead of $X_t'$.

\textbf{Total Loss}
The total loss for the Generator network $G$ amounts to:
\begin{equation}
    L_G = L_{CY} w_{CY} + L_I w_I + L_A  w_A + L_{CO} w_{CO}
\end{equation}
Where $w_{CY}$, $w_{I}$, $w_{A}$, and $w_{CO}$ are weights to balance the contribution of the different losses to the total generator loss.

\section{Experimental Setup}
\label{sec:experimental_setup}

\subsection{Datasets}
\label{ssc:datasets}

\textbf{MIT5K~\cite{fivek}.} 
The dataset is composed of 5,000 high resolution images that are retouched by five experts performing global and local adjustments. We follow the experimental setting of \cite{DPE} and use 2,250 images and their retouched adaptations for training the supervised models. 
The test set contains 500 images. The remaining 2,250 images are reserved for the target domain in the unsupervised setting. 
For the multitask experiments data of all five experts is used, thus increasing the amount of train and test images to 11,250 and 2,500, respectively.

\textbf{DPED~\cite{ignatov2017dslr}.} 
DPED contains paired data of scenes captures with three smartphone cameras, \ie, iPhone 3GS, BlackBerry Passport and Sony Xperia Z, and a professional quality DSLR, Cannon 70D. 
In total the dataset contains 22K full resolution photos. To get paired samples suited for supervised training, the authors generated aligned patches of $100\times 100$ pixels. There are 139K, 160K and 162K training patches for BlackBerry, iPhone and Sony smartphones respectively. Correspondingly, there are 2.4K, 4.4K and 2.5K test patches.

\begin{figure}[bh]
\centering
\subfigure{
\begin{minipage}[t]{0.1\linewidth}
\centering
\includegraphics[width=1\textwidth]{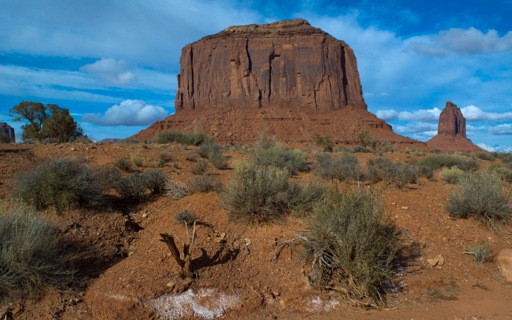}
\end{minipage}
\begin{minipage}[t]{0.1\linewidth}
\centering
\includegraphics[width=1\textwidth]{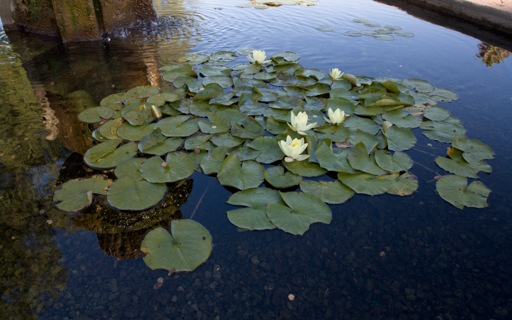}
{\footnotesize Normal}
\end{minipage}
\begin{minipage}[t]{0.1\linewidth}
\centering
\includegraphics[width=1\textwidth]{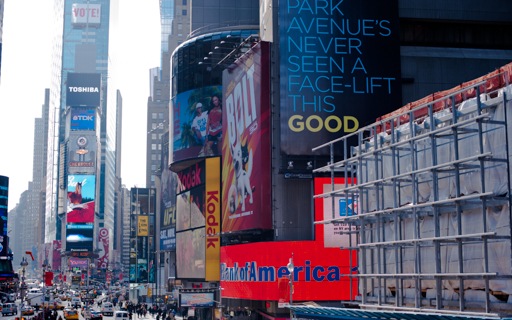}
\end{minipage}~
\begin{minipage}[t]{0.1\linewidth}
\centering
\includegraphics[width=1\textwidth]{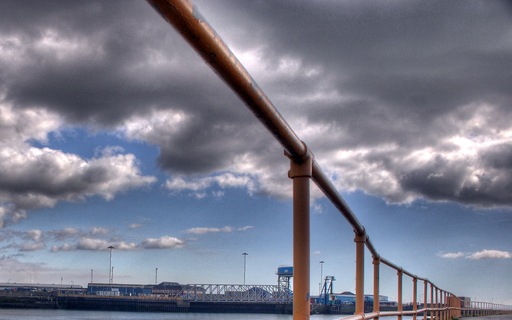}
\end{minipage}
\begin{minipage}[t]{0.1\linewidth}
\centering
\includegraphics[width=1\textwidth]{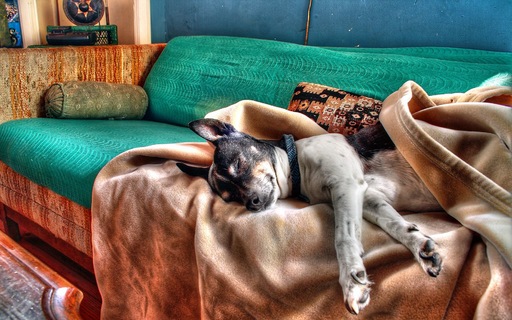}
{\footnotesize HDR}
\end{minipage}
\begin{minipage}[t]{0.1\linewidth}
\centering
\includegraphics[width=1\textwidth]{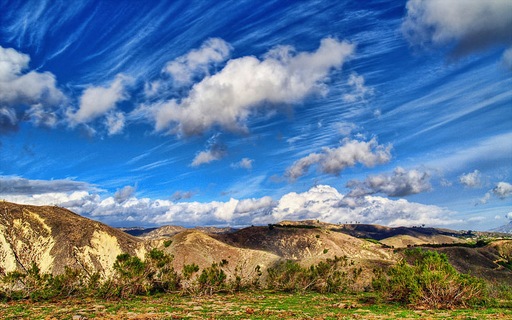}
\end{minipage}~
\begin{minipage}[t]{0.1\linewidth}
\centering
\includegraphics[width=1\textwidth]{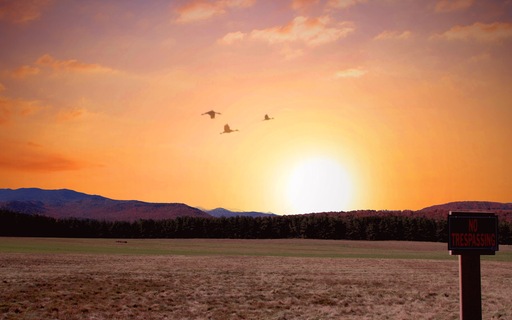}
\end{minipage}
\begin{minipage}[t]{0.1\linewidth}
\centering
\includegraphics[width=1\textwidth]{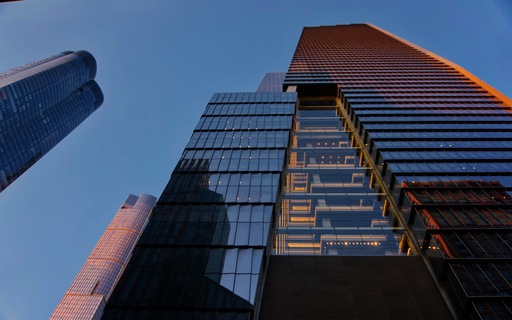}
{\footnotesize Sunset}
\end{minipage}
\begin{minipage}[t]{0.1\linewidth}
\centering
\includegraphics[width=1\textwidth]{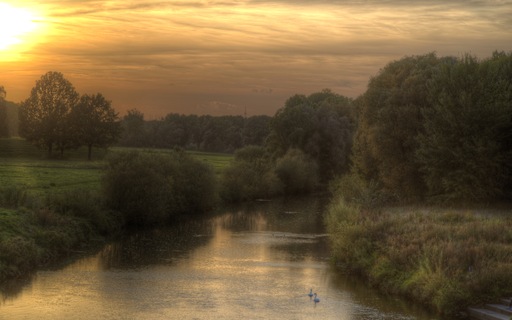}
\end{minipage}
}
\caption{Visual examples of the images in different subset of the constructed Flickr dataset.}
\label{fig:flickrdataset}
\end{figure}

\textbf{Flickr Multi-style.}
To test our approach on a dataset in the wild, we created a small multi class dataset. There are three style classes.
\textit{1) Normal}, which consists of a subset of 600 images from the MIT5K dataset.
\textit{2) Sunset}, which consists of 600 images with the tag sunset that are collected from Flickr and selected acording to Flickr's interestness score.
\textit{3) HDR}, which is a subset of 594 of the HDR-Flickr dataset from \cite{DPE} that are also collected using the Flickr API.
 Some visual examples of the constructed dataset can be found in Fig. \ref{fig:flickrdataset}.

\subsection{Performance measures}
In our experiments we use the standard Peak Signal to Noise Ratio (PSNR) and Structual Similarity Index (SSIM) to measure the fidelity of the enhanced images towards the ground truth / target reference images. Complementary we use LPIPS~\cite{LPIPS2018}, a learned perceptual metric meant to approximate the human perceptual similarity judgements.

\subsection{Implementation Details}
We use the TensorFlow framework for all experiments. For the supervised experiments on the MIT5K dataset~\cite{fivek}, we follow the setup used in \cite{DPE}. This includes using the exact train and test split, in order to get comparable results. The size of the input layer is fixed at $512\times 512$ pixels during training. Images are resized for the longer edge to equal 512 pixels prior to training. To match the input of the network the images are zero padded.
In all supervised experiments we use the following cost function, which maximizes the PSNR: $L_G = -log_{10}(MSE(I_s, I_t))$.
The models are trained for 100k iterations with batch size 4 and learning rate 1e-4 on MIT5K dataset, while on DPED dataset the models are trained for 150k iterations with batch size 50 and learning rate 1e-4.

\section{Results on Supervised Image Enhancement}
\label{sec:supervised_results}

In this section, we evaluate our method for supervised image enhancement. For more visual results we refer to the supplementary material.

\begin{table}[bp!]
\centering
\begin{tabular}{l|ccr}
\hline
\textbf{Method} & \textbf{PSNR} & \textbf{SSIM} & \textbf{Para.} \\
\hline\hline
DPED ~\cite{ignatov2017dslr} & 21.76 & 0.871 & 401k \\
DPE ~\cite{DPE} & 23.80 & 0.900 & 5019k \\
UPEDIE ~\cite{Wang_2019_CVPR} & 23.04 & 0.893 & -\\
\hline\hline
SGN2 & 20.98 &0.863 & 769k \\
GSGN w/o global feat. \& IN & 21.19 & 0.865& 325k \\
GSGN w/o IN & 23.74 & 0.900& 338k \\
\hline
\hline
GSGN (Ours)  & \textbf{24.16} & \textbf{0.905} & 339k \\
\hline
\end{tabular}
\caption{Comparison between state-of-the-art and our GSGN with different configurations on MIT5K dataset (Expert C). We refer the reader to the latest \textbf{corrected} version of the UPEDIE paper~\cite{Wang_2019_CVPR}.}
\label{tab:ablation_study_and_other_work}
\end{table}

\subsection{Single task enhancement}
\label{ssc:single_task}

To justify our design choices as well as to compare with other state-of-the-art approaches, we firstly conduct experiments on the supervised single task enhancement setting.
We follow the experimental setting of \cite{DPE} and 
train different models to approximate the retouched results by expert C in the MIT5K dataset~\cite{fivek}.

\begin{figure*}[t!]
\centering
\setlength{\tabcolsep}{1pt}
\resizebox{\linewidth}{!}
{
\begin{tabular}{cccc}
    \includegraphics[width=0.29\linewidth]{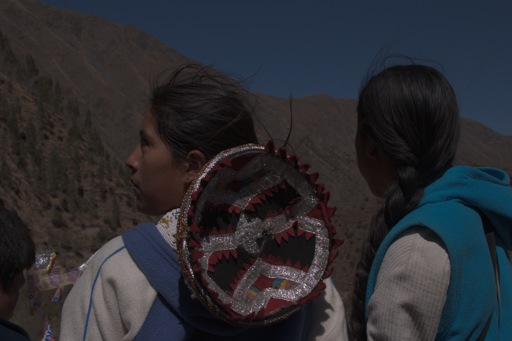}&
    \includegraphics[width=0.29\linewidth]{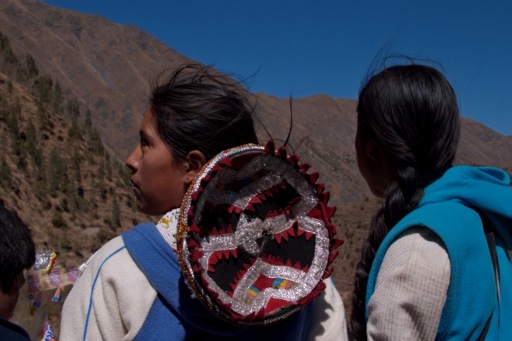}&
    \includegraphics[width=0.29\linewidth]{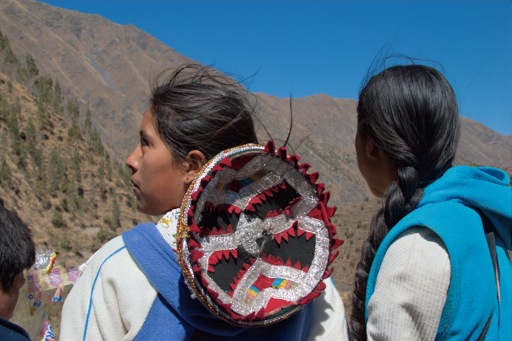}&
    \includegraphics[width=0.29\linewidth]{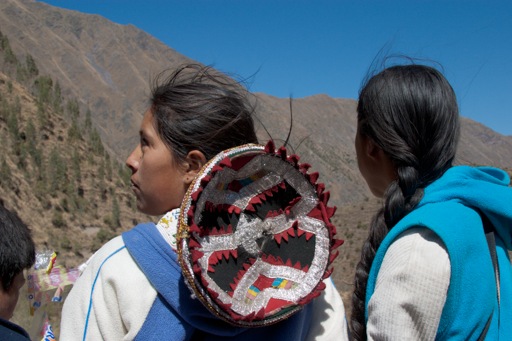}\\
    \includegraphics[width=0.29\linewidth]{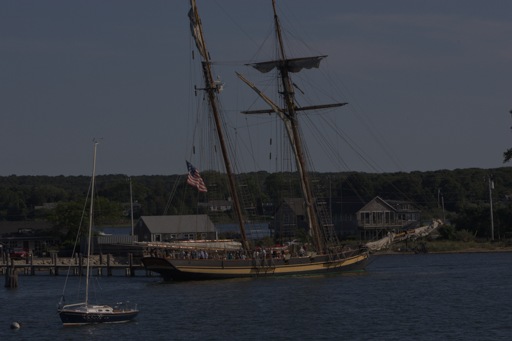}&
    \includegraphics[width=0.29\linewidth]{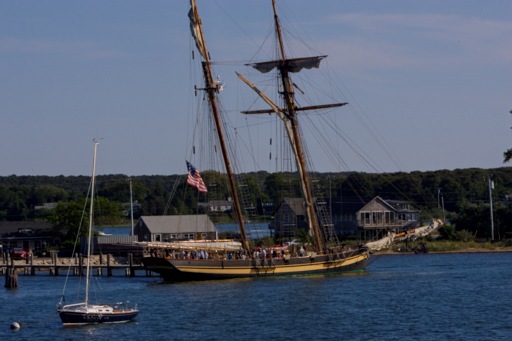}&
    \includegraphics[width=0.29\linewidth]{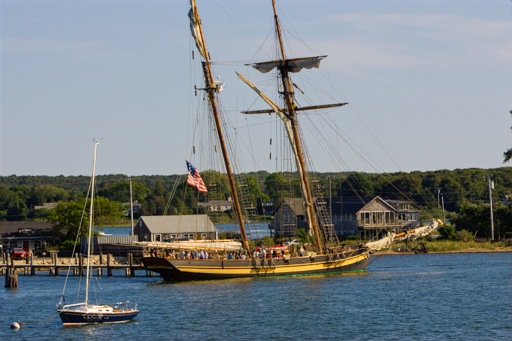}&
    \includegraphics[width=0.29\linewidth]{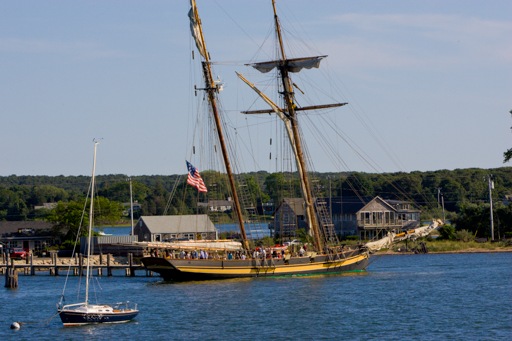}\\
    \includegraphics[width=0.29\linewidth]{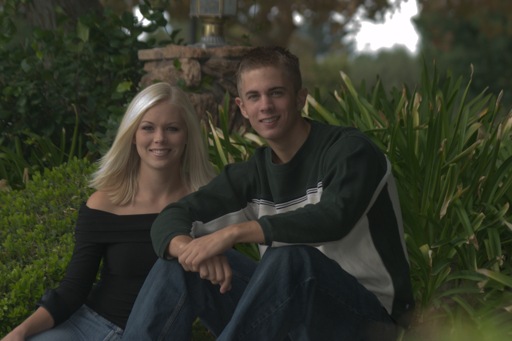}&
    \includegraphics[width=0.29\linewidth]{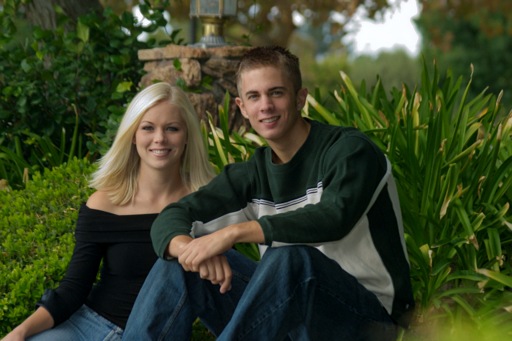}&
    \includegraphics[width=0.29\linewidth]{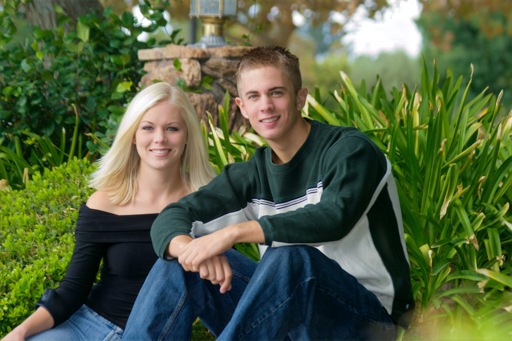}&
    \includegraphics[width=0.29\linewidth]{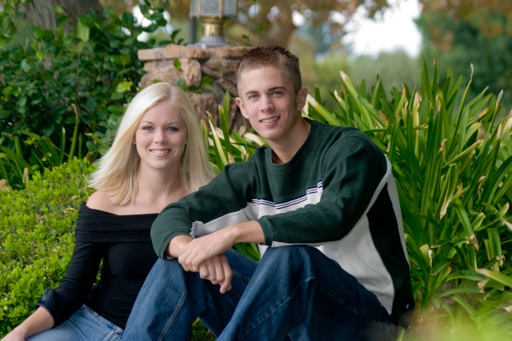}\\
    input & DPE SL & Ours SL & label\\
\end{tabular}
}
\caption{Visual comparison of learned supervised mappings of our GSGN network against the DPE~\cite{DPE} method on the test set of the MIT5K dataset.}
\label{fig:visual_comparison_supervised}
\end{figure*}

Concretely, we compare the proposed GSGN network with the original 2 level SGN~\cite{SGN_ICCV}, denoted by SGN2; and the SGN2 with reduced number filter channel numbers, denoted by GSGN w/o global feat. \& instance normalization; SGN2 with reduced number filter channel numbers and global feature branch, denoted by GSGN w/o instance normalization; and our final GSGN model.
In addition to our ablation study, we also compare GSGN with three recently proposed architectures that were designed for this task.
The comparison methods include DPED~\cite{ignatov2017dslr}, DPE~\cite{DPE} and UPEDIE~\cite{Wang_2019_CVPR}.
DPED~\cite{ignatov2017dslr} uses a fully convolutional ResNet for image enhancement.
DPE~\cite{DPE} used U-Net~\cite{U-Net} as a backbone and augmented it with global features. 
UPEDIE~\cite{Wang_2019_CVPR} adopts an alternative approach by learning an illumination map from an encoder network. 
Table~\ref{tab:ablation_study_and_other_work} shows the average PSNR and SSIM of the different approaches, evaluated on the 500 test images of the MIT5K dataset. 
Figure~\ref{fig:visual_comparison_supervised} shows visual results of the proposed GSGN network.
For the results reported in \cite{DPE} we were not able to reproduce their high PSNR and SSIM, to be fair we still use the original numbers they claim in the paper. The results show, that our network not only outperforms its competitors by at least 0.36 dB, but also uses considerably less parameters to achieve these results.
The results in table~\ref{tab:ablation_study_and_other_work} validate the effectiveness of the proposed GSGN architecture.
GSGN achieved higher PSNR and SSIM while using fewer parameters than current state-of-the-art methods.

\subsection{Multi-task Enhancement: One-to-Many}
\label{ssc:multi_task}
The fact that the Adobe MIT-5K dataset contains five retouched versions of each input image, makes it ideally suited to be used as a base for multitask learning in a supervised one-to-many setting.
In order to demonstrate the feasibility of our task adaptive GSGN model, we train models to capture all five input to output mappings of the MIT5K dataset simultaneously. 
We consider the following three experimental settings. Results are found in table~\ref{tab:comparison_multitask_mit}.

\begin{figure*}[t!]
\centering
\setlength{\tabcolsep}{1pt}
\resizebox{\linewidth}{!}
{
\begin{tabular}{cccccc}
    \includegraphics[width=0.19\linewidth]{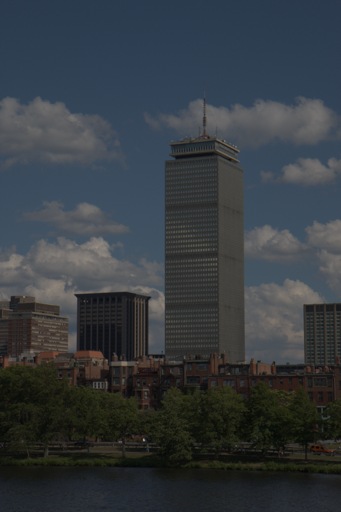}&
    \includegraphics[width=0.19\linewidth]{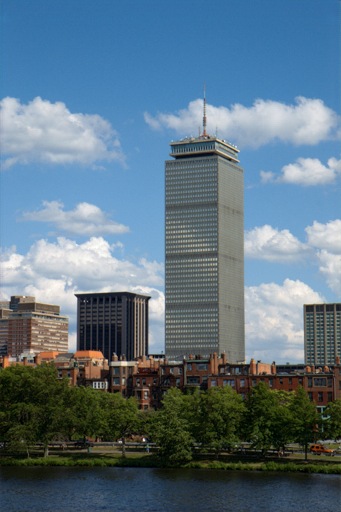}&
    \includegraphics[width=0.19\linewidth]{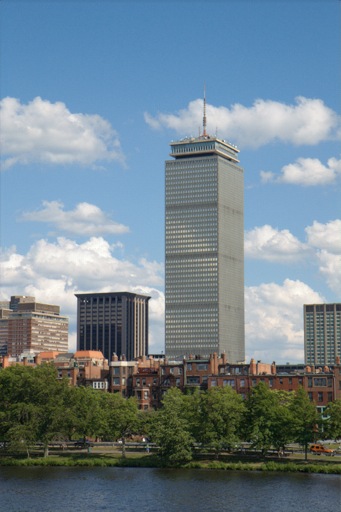}&
    \includegraphics[width=0.19\linewidth]{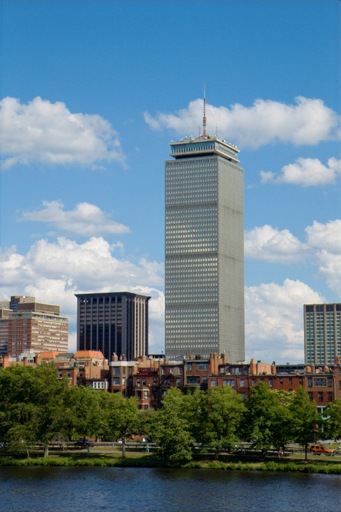}&
    \includegraphics[width=0.19\linewidth]{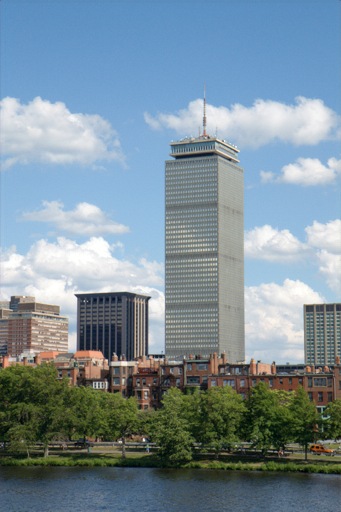}&
    \includegraphics[width=0.19\linewidth]{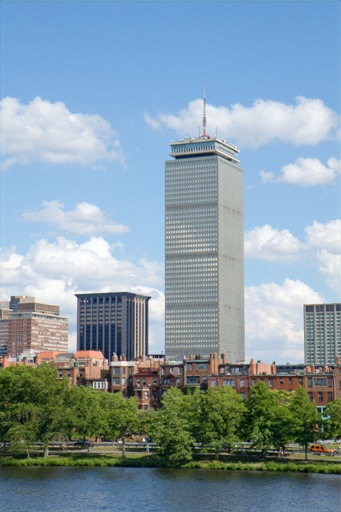}\\
    input & learned A & learned B & learned C & learned D & learned E\\
    &
    \includegraphics[width=0.19\linewidth]{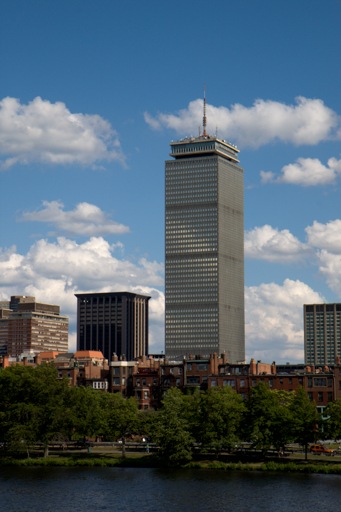}&
    \includegraphics[width=0.19\linewidth]{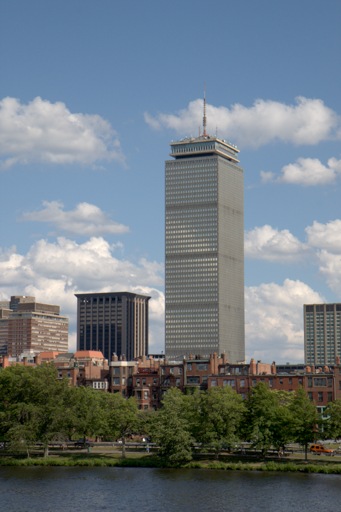}&
    \includegraphics[width=0.19\linewidth]{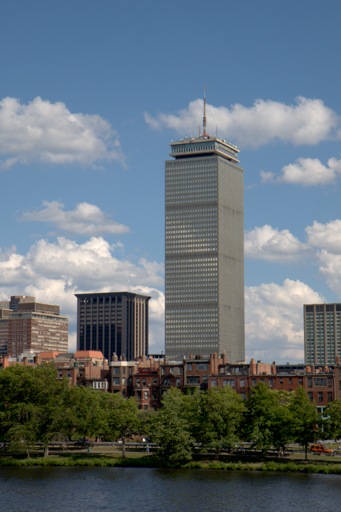}&
    \includegraphics[width=0.19\linewidth]{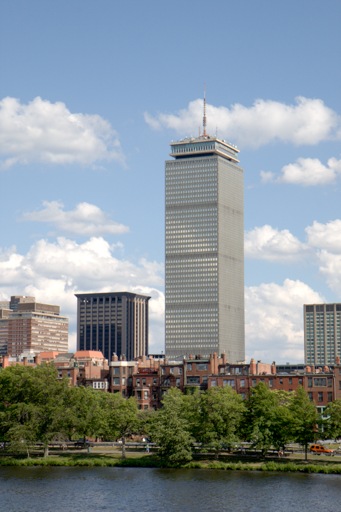}&
    \includegraphics[width=0.19\linewidth]{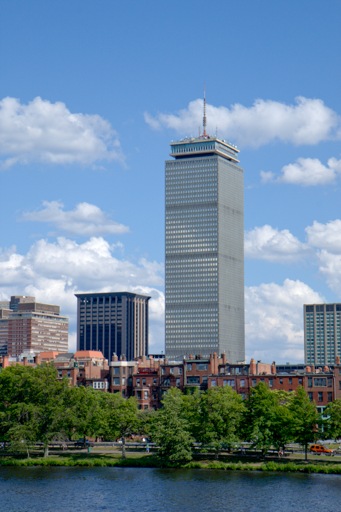}\\
    & expert A & expert B & expert C & expert D & expert E\\
\end{tabular}
}
\caption{Comparison of learned supervised mappings by our MT-GSGN model and corresponding expert labels. }
\label{fig:example_results}
\end{figure*}

\paragraph{Single Task Models.}
We train our GSGN model on the task of learning all input to output mappings separately for all five experts A to E in a traditional one-to-one setting.
The averaged PSNR over all five experiments is 23.97 dB. It is worth noting that there is quite a large difference of up to 3.82 dB PSNR between the different experts. This gives interesting insights into the dataset. On one side this could mean that certain experts were more consistent when applying the mapping. On the other hand, this could also indicate that some experts used more complex adjustments than others, which made it harder for the network to learn these mappings.

\paragraph{Single Model for All Tasks.}
The GSGN model is trained on data of all five task simultaneously in a traditional one-to-many setting but without a constant latent embedding so it can not explicitly discriminate between the different tasks.
This effectively increases the amount of training data by a factor of five. However, the network is only able to learn a average mapping and cannot take the specific style of each expert into account. While performance gains come from being trained on larger amounts of data, the lack of being able to differentiate between the different experts causes the performance to decrease, this results in a lower average PSNR of 23.30 dB.
\paragraph{Multi-tasking with a Task Adaptive Model.}
We use our task adaptive GSGN model to learn all five tasks simultaneously, in a one-to-many setting.
This combines the advantages of the previous two settings. The resulting model reaches the highest average PSNR of 24.32 dB and also outperforms the other settings in terms of SSIM, MSSIM and LPIPS~\cite{LPIPS2018}. This not only shows that the model is able to efficiently use the labels to learn multiple mappings simultaneously, but also that by taking advantage of the larger amounts of data available in a multitask setting, the model is able to outperform models that were trained for a single task.
Figure~\ref{fig:example_results} shows visual results of the MT-GSGN model and the corresponding expert labels.

\begin{table}
\centering
\resizebox{\linewidth}{!}
{
\begin{tabular}{c|ccc|ccc|ccc}
\hline
{\multirow{2}{*}{\diagbox{Tasks}{Methods}}}
& \multicolumn{3}{c|}{Task-Specific Models}
& \multicolumn{3}{c|}{Single Model for All Tasks} 
& \multicolumn{3}{c}{Multi-tasking (MT-GSGN)}\\
\cline{2-10}
 & PSNR & SSIM  & LPIPS&PSNR & SSIM  & LPIPS&PSNR & SSIM  & LPIPS\\
\hline\hline
\multicolumn{10}{c}{Supervised enhancement}\\
\hline
Raw $\rightarrow$ A & 22.30 & 0.876  & 0.0743 & 21.28 & 0.862 & 0.080 & 22.51 & 0.879 & 0.072\\
Raw $\rightarrow$ B & 26.12 & 0.949  & 0.0443 & 25.48 & 0.939 & 0.049 & 26.53 & 0.952 & 0.043 \\
Raw $\rightarrow$ C & 24.16 & 0.905 & 0.0610 & 23.71 & 0.896 & 0.069 & 24.44 & 0.904 & 0.063 \\
Raw $\rightarrow$ D & 23.02 & 0.903  & 0.0621 & 23.14 &0.905 & 0.063 & 23.27 & 0.906 & 0.060   \\
Raw $\rightarrow$ E & 24.23 & 0.924  & 0.0571 & 22.92 & 0.911 & 0.067 & 24.86 & 0.928 & 0.055 \\
\hline\hline
Avg. & 23.97 & 0.911  & 0.0598&23.30 & 0.903  & 0.066&\textbf{24.32} & \textbf{0.914} & \textbf{0.059} \\
\hline
\multicolumn{10}{c}{Unsupervised enhancement}\\
\hline\hline
Raw $\rightarrow$ A & 18.95 & 0.771 & 0.212 & 16.63 & 0.705 & 0.289 & 19.99 & 0.824 & 0.123\\
Raw $\rightarrow$ B & 20.49 & 0.840 & 0.204 & 18.23 & 0.809 & 0.255 & 22.30 & 0.908 & 0.097 \\
Raw $\rightarrow$ C & 19.16 & 0.779 & 0.228 & 16.76 & 0.707 & 0.304 & 20.81 & 0.839 & 0.120 \\
Raw $\rightarrow$ D & 17.92 & 0.767 & 0.246 & 16.46 & 0.743 & 0.287 & 20.36 & 0.866 & 0.111  \\
Raw $\rightarrow$ E & 17.75 & 0.758 & 0.271 & 16.03	& 0.730 & 0.311 & 19.89 & 0.859 & 0.124 \\
\hline\hline
Avg. & 18.85 & 0.783 & 0.232 & 16.82 & 0.739 & 0.289 & \textbf{20.67} & \textbf{0.859} & \textbf{0.115} \\
\hline
\end{tabular}
}
\caption{Enhancement one-to-one and \textbf{one-to-many} results on MIT5K.}
\label{tab:comparison_multitask_mit}
\end{table}

\subsection{Multi-task Enhancement: Many-to-One}
We further validate our model on a many-to-one multi-task mapping on DPED~\cite{ignatov2017dslr} dataset and train models to enhance the images from the iPhone, Sony and Blackberry cameras to the DSLR camera domain. We follow the same three step approach described in the previous section, and compare the average performance of the single task model against the multi-task model. 
It is worth noting that in this case a weighted average is needed because the number of images in the training sets for each phone is different.

\begin{table}
\centering
\begin{tabular}{l|ccc|ccc}
\hline
&\multicolumn{3}{c|}{Supervised}&\multicolumn{3}{c}{Unsupervised}\\
Method & PSNR & SSIM & LPIPS & PSNR & SSIM & LPIPS\\
\hline\hline
Iphone & 22.94 & 0.819 & 0.142 & 19.47 & 0.648 & 0.264 \\
Sony & 24.46 & 0.877 & 0.103 & 23.21 & 0.821 & 0.142 \\
Blackberry & 23.18 & 0.842 & 0.108 & 19.44  & 0.640 & 0.225 \\
\hline
Weighted avg. & 23.41 & \textbf{0.841} & \textbf{0.122}  & 20.48 & 0.640 & 0.225 \\
\hline\hline
Single Model (all) & 23.35 &	0.836 &	0.129  & 20.22 &	0.685 &	0.238 \\
Multi-tasking & \textbf{23.69} & 0.839 & 0.128 & \textbf{20.69} & \textbf{0.749} & \textbf{0.209} \\
\hline
\end{tabular}
\caption{Enhancement one-to-one and \textbf{many-to-one} results on DPED.}
\label{tab:comparison_multitask_manytoone}
\end{table}

Table~\ref{tab:comparison_multitask_manytoone} shows the evaluation results for this experiment. In this case the multi-task model outperforms its single task counterparts, in terms of PSNR, but not in terms of SSIM and LPIPS. This is likely due to the fact that the model is trained to optimize PSNR but training patches coming from different sources can never be perfectly aligned due to nonlinear distortions coming from different lenses and sensors, as well as perspective distortions. The authors of \cite{ignatov2017dslr} note that there could be shift of up to 5 pixels between source and target images. This makes the problem formulation ill posed for supervised learning and suggests that this dataset is better suited for weakly supervised learning methods.

\section{Results on Unsupervised Image Enhancement}
\paragraph{Implementation Details}
As described in section~\ref{sec:proposed_methods}, we follow a two cycle GAN approach inspired by \cite{ACGAN2018}, where an additional conditional loss is used to provide task specific gradients during training.
A discriminator architecture similar to the one described in \cite{DPE} is used.
We train the critic 30 times more than the generator for the MIT5K and Flickr Multi-style datasets and 40 times more often for the DPED dataset.

\paragraph{Multi-task Enhancement: One-to-Many}
We evaluate our model on MIT5K dataset, using our weakly supervised GAN architecture. Analog to the supervised experiments, we perform the same three experimental settings. Table~\ref{tab:comparison_multitask_mit} shows the validation results. Our multi-task learning approach with task adaptive GSGN again outperforms, our GSGN network by taking advantage of shared feature representations.

\begin{figure*}[t!]
\centering
\setlength{\tabcolsep}{1pt}
\resizebox{\linewidth}{!}
{
\begin{tabular}{cccccc}
    \includegraphics[width=0.19\linewidth]{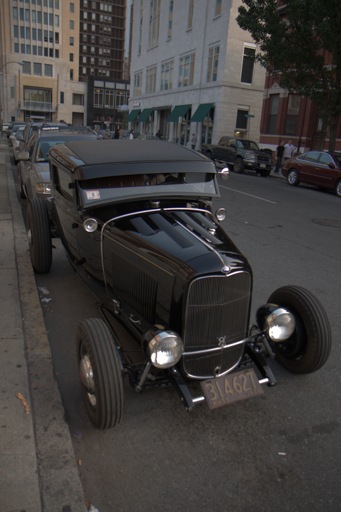}&
    \includegraphics[width=0.19\linewidth]{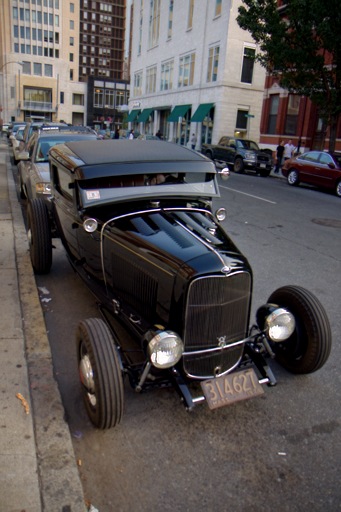}&
    \includegraphics[width=0.19\linewidth]{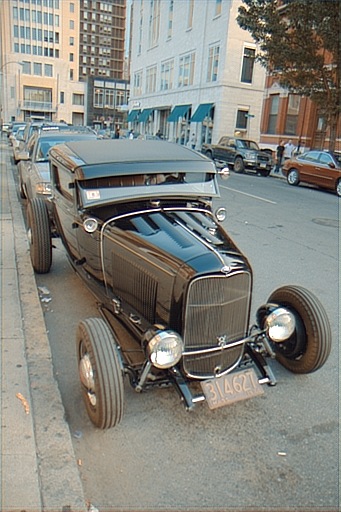}&
    \includegraphics[width=0.19\linewidth]{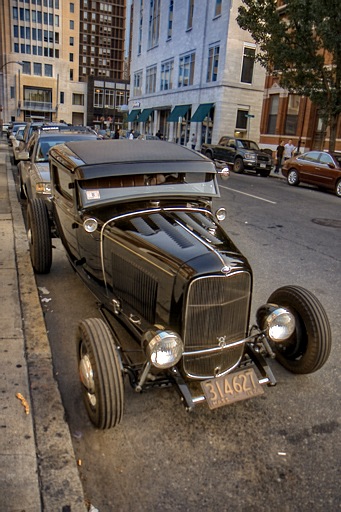}&
    \includegraphics[width=0.19\linewidth]{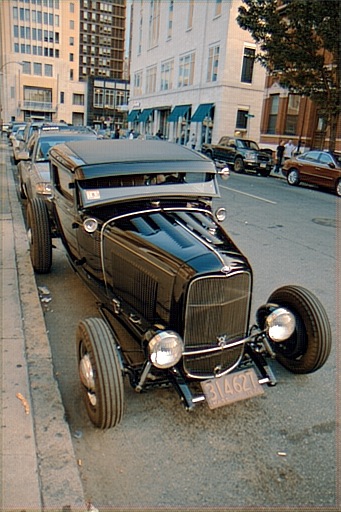}&
    \includegraphics[width=0.19\linewidth]{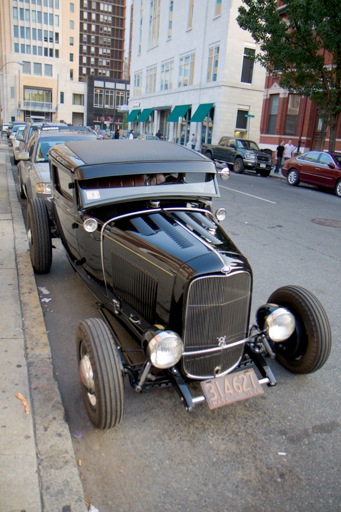}\\
    \includegraphics[width=0.19\linewidth]{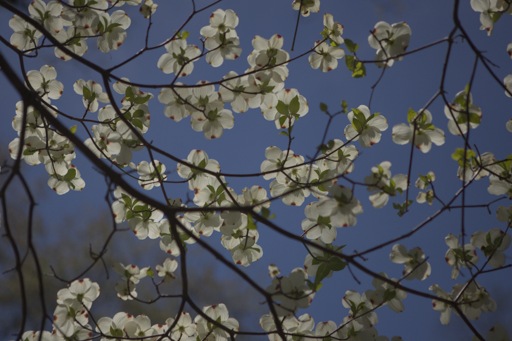}&
    \includegraphics[width=0.19\linewidth]{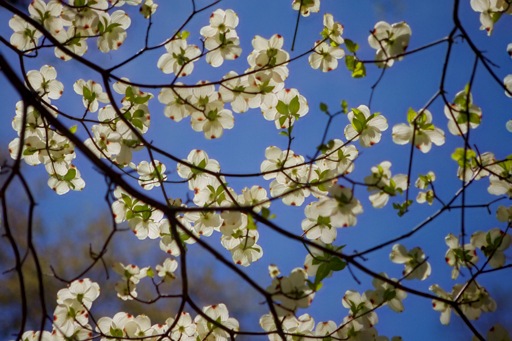}&
    \includegraphics[width=0.19\linewidth]{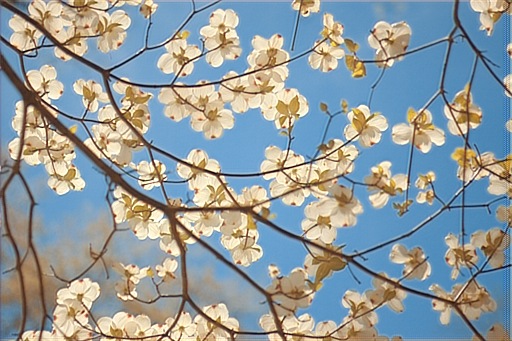}&
    \includegraphics[width=0.19\linewidth]{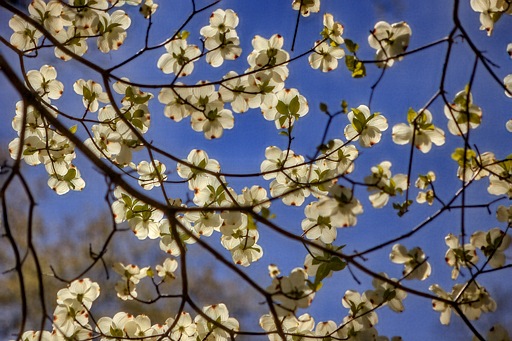}&
    \includegraphics[width=0.19\linewidth]{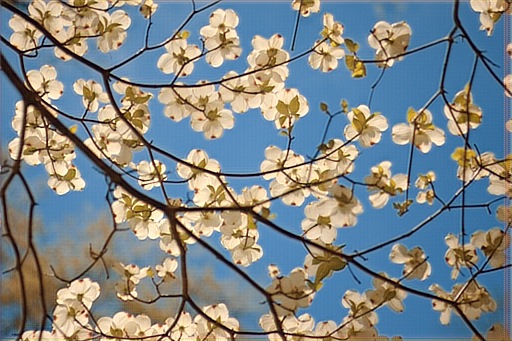}&
    \includegraphics[width=0.19\linewidth]{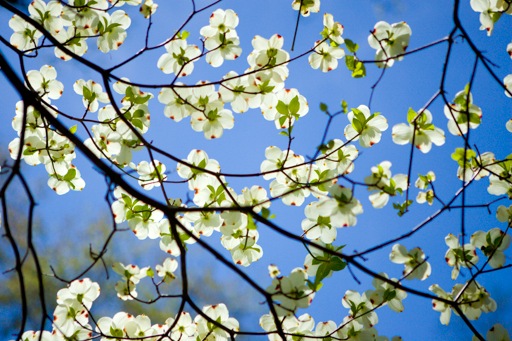}\\
    \includegraphics[width=0.19\linewidth]{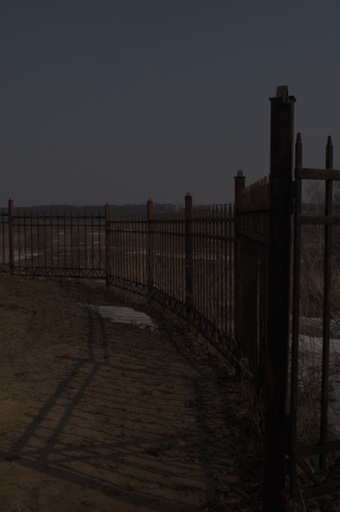}&
    \includegraphics[width=0.19\linewidth]{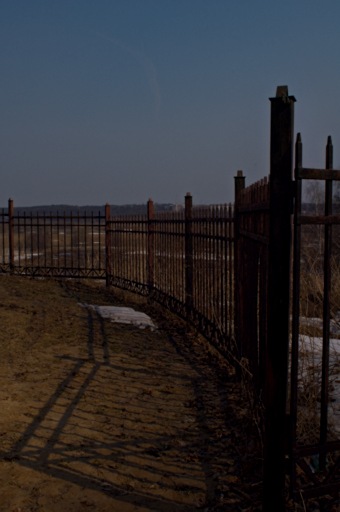}&
    \includegraphics[width=0.19\linewidth]{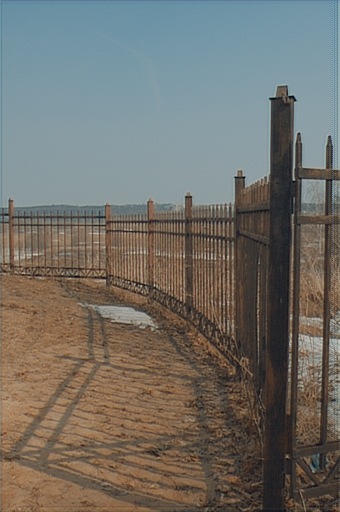}&
    \includegraphics[width=0.19\linewidth]{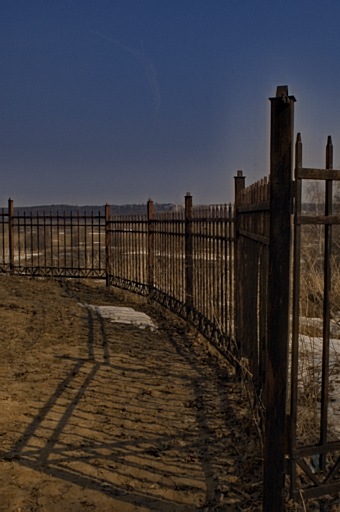}&
    \includegraphics[width=0.19\linewidth]{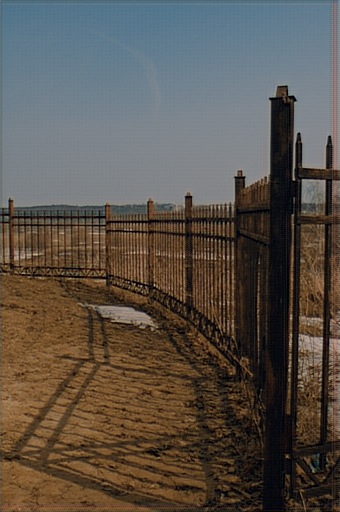}&
    \includegraphics[width=0.19\linewidth]{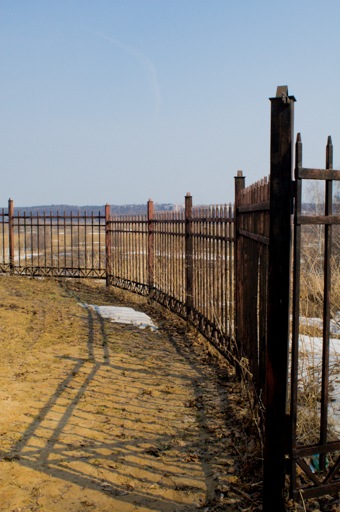}\\
    input & DPE UL & Ours UL & DPE HDR & Ours HDR & label\\
\end{tabular}
}
\caption{Visual comparison of learned unsupervised mappings by our Task Adaptive GSGN model trained on the small Flickr Multi-style dataset, against the DPE~\cite{DPE} method. Note that our model learns both mapppings in a single training while DPE requires separate networks and multiple trainings.
}
\label{fig:visual_comparison_flickr}
\end{figure*}

\paragraph{Multi-task Enhancement: Many-to-One}
We validate the proposed algorithm on the many-to-one task on the DPED dataset. In contrast to the one-to-many setting, this setting does not require a conditional loss on the generated images, which simplifies the training. Table ~\ref{tab:comparison_multitask_manytoone} shows the evaluation results.  
\paragraph{Flickr Multi-style Dataset}
\label{ssc:fliker_results}
\begin{figure*}[h!]
\centering
\subfigure{
\begin{minipage}[t]{0.14\linewidth}
\centering
\includegraphics[width=1\textwidth]{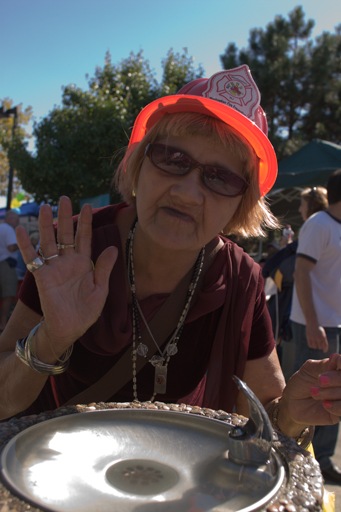}
{\footnotesize  (a)}
\end{minipage}
\begin{minipage}[t]{0.14\linewidth}
\centering
\includegraphics[width=1\textwidth]{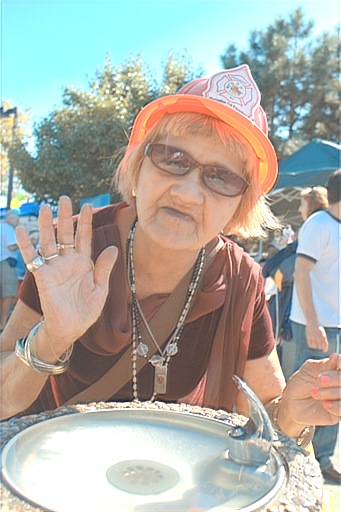}
{\footnotesize  (b)}
\end{minipage}
\begin{minipage}[t]{0.14\linewidth}
\centering
\includegraphics[width=1\textwidth]{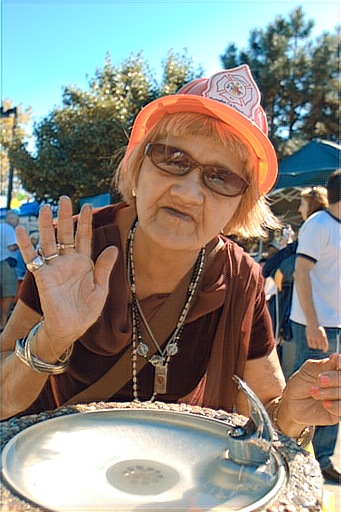}
{\footnotesize  (c)}
\end{minipage}
\begin{minipage}[t]{0.14\linewidth}
\centering
\includegraphics[width=1\textwidth]{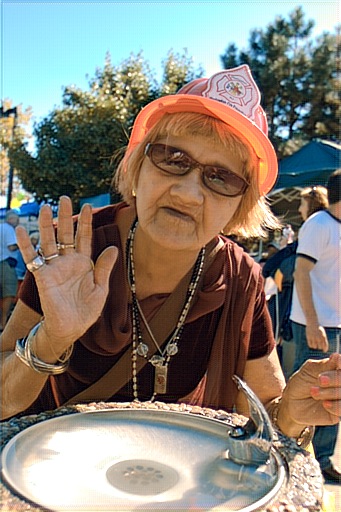}
{\footnotesize  (d)}
\end{minipage}
\begin{minipage}[t]{0.14\linewidth}
\centering
\includegraphics[width=1\textwidth]{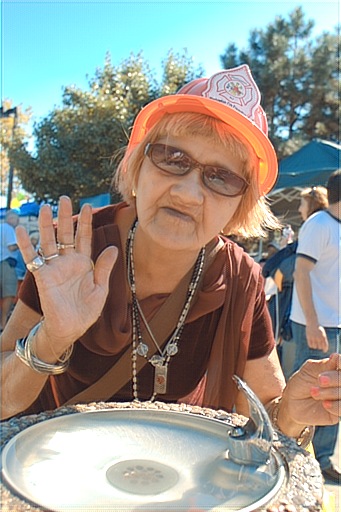}
{\footnotesize  (e)}
\end{minipage}
\begin{minipage}[t]{0.14\linewidth}
\centering
\includegraphics[width=1\textwidth]{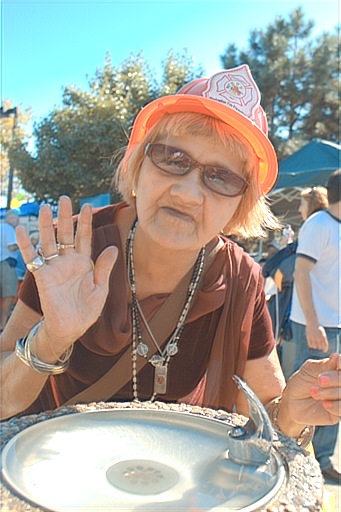}
{\footnotesize  (f)}
\end{minipage}
}
\caption{Visual examples of enhanced images from MT-GSGN trained on the Flickr dataset. (a) Input, (b) Normal, (c) Interpolation between HDR and Sunset, (d) HDR, (e) Sunset, (f) Interpolation between Normal and HDR. We refer to 
sec.~\ref{ssc:fliker_results} for details.}
\label{fig:flickr_enhanced_examples}
\end{figure*}
Finally, we evaluate our MT-GSGN model on a dataset on the wild. For this purpose we train it on our Flickr dataset on the task of learning a one-to-many mapping with three styles to learn, normal, sunset and hdr.
Figure~\ref{fig:flickr_enhanced_examples} shows visual examples of the learned mappings.  Note that the model is also able to interpolate between styles, even though it was not explicitly trained for this task.
Figure~\ref{fig:visual_comparison_flickr} shows a visual comparison of the learned mappings against the state of the art.
For more visual results we refer to the supplementary material.

\section{Conclusion}
In this paper, we have investigated and proposed a flexible example-based image enhancement method, that uses a task adaptive global feature self-guided network. First, we proposed a novel network architecture capable to outperform existing methods on the task of supervised image to image mapping while requiring much fewer parameters. Second, we demonstrated that by using multi-task learning we benefit from shared representation and achieve higher performance compared to learning one to one mappings separately. Third, for both supervised and unsupervised settings our experimental results validate the effectiveness of our multi-task learning for image enhancement in one-to-many and many-to-one settings. To the best of our knowledge this is the first successful work in applying multi task learning to the challenging image enhancement problem.

\section*{Acknowledgements}
This work was partly supported by ETH Zurich General Fund (OK), by a Huawei project and by Amazon AWS and Nvidia grants.

\clearpage
\bibliographystyle{splncs04}
\bibliography{egbib}

\clearpage
\onecolumn

\appendix

\setcounter{table}{0}
\renewcommand{\thetable}{A\arabic{table}}
\setcounter{figure}{0}
\renewcommand{\thefigure}{A\arabic{figure}}

\section{Flexible Example-based Image Enhancement with Task Adaptive Global Feature Self-Guided Network -- Supplementary}\vspace{2em}

\subsection{Supervised learning - Visual comparison against the State-of-the-art}
\begin{figure*}[h!]
\centering
\setlength{\tabcolsep}{1pt}
\resizebox{\linewidth}{!}
{
\begin{tabular}{cccc}
    \includegraphics[width=0.29\linewidth]{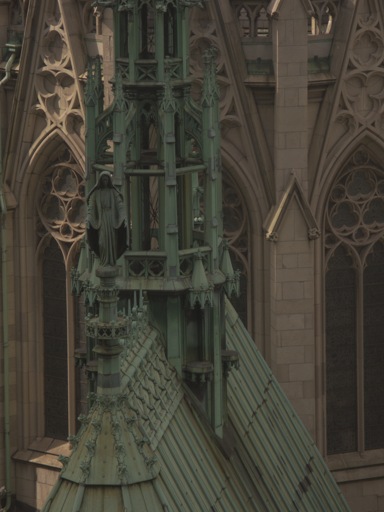}&
    \includegraphics[width=0.29\linewidth]{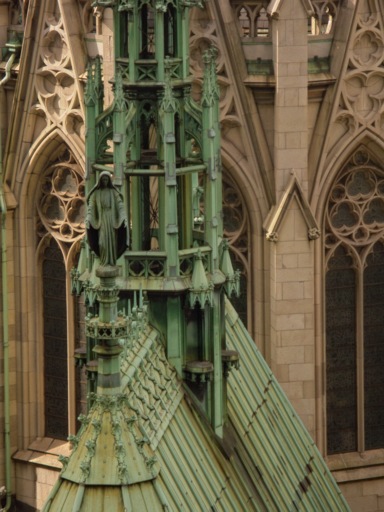}&
    \includegraphics[width=0.29\linewidth]{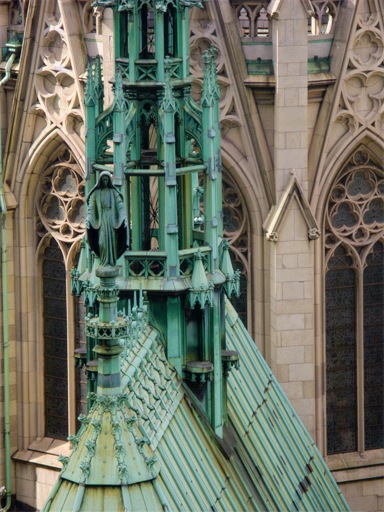}&
    \includegraphics[width=0.29\linewidth]{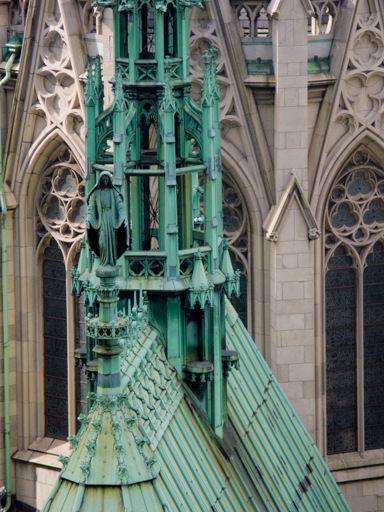}\\
    \includegraphics[width=0.29\linewidth]{VisualComparisonMIT5K_paired/a0503_source.jpg}&
    \includegraphics[width=0.29\linewidth]{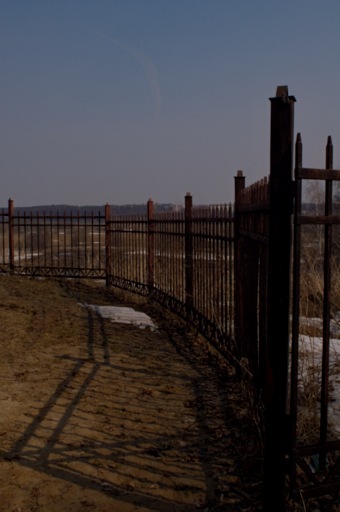}&
    \includegraphics[width=0.29\linewidth]{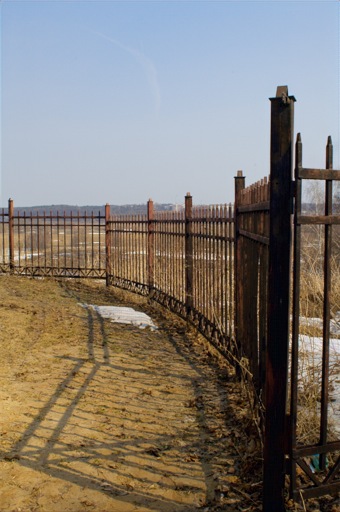}&
    \includegraphics[width=0.29\linewidth]{VisualComparisonMIT5K_paired/a0503_target.jpg}\\
    \includegraphics[width=0.29\linewidth]{VisualComparisonMIT5K_paired/a1581_source.jpg}&
    \includegraphics[width=0.29\linewidth]{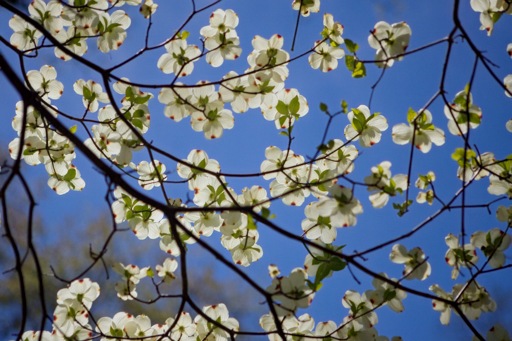}&
    \includegraphics[width=0.29\linewidth]{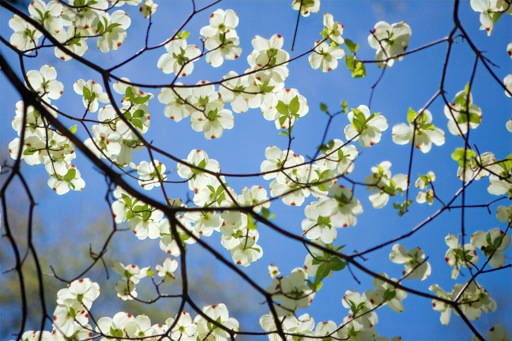}&
    \includegraphics[width=0.29\linewidth]{VisualComparisonMIT5K_paired/a1581_target.jpg}\\
    input & DPE SL & Ours SL & label\\
\end{tabular}
}
\vspace{-0.2cm}
\caption{Visual comparison of learned supervised mappings of our GSGN network against the DPE~\cite{DPE} method on images of the test set of the MIT5K~\cite{fivek} dataset (expert~C).}
\label{fig:supplement_visual_comparison_supervised}
\end{figure*}

\begin{figure*}[h!]
\centering
\setlength{\tabcolsep}{1pt}
\resizebox{\linewidth}{!}
{
\begin{tabular}{cccc}
    \includegraphics[width=0.29\linewidth]{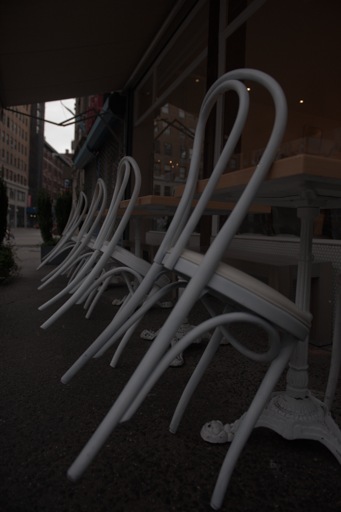}&
    \includegraphics[width=0.29\linewidth]{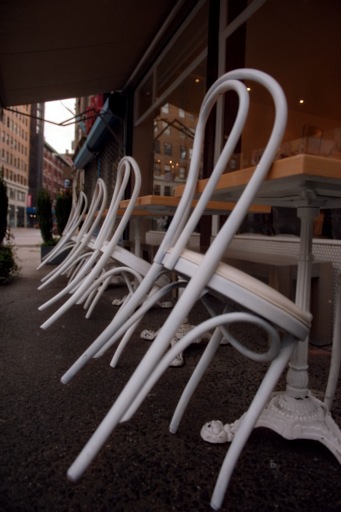}&
    \includegraphics[width=0.29\linewidth]{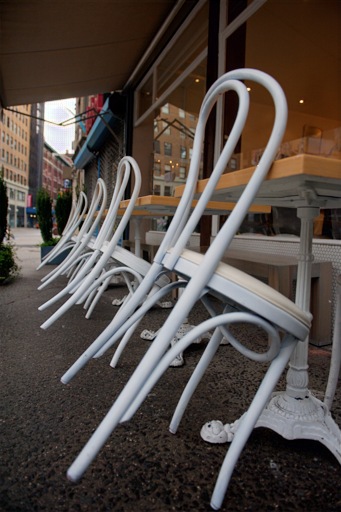}&
    \includegraphics[width=0.29\linewidth]{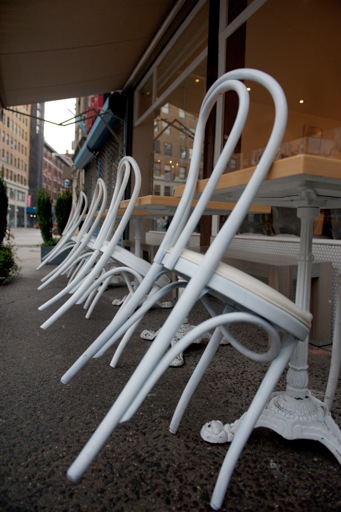}\\
    \includegraphics[width=0.29\linewidth]{VisualComparisonMIT5K_paired/a0688_source.jpg}&
    \includegraphics[width=0.29\linewidth]{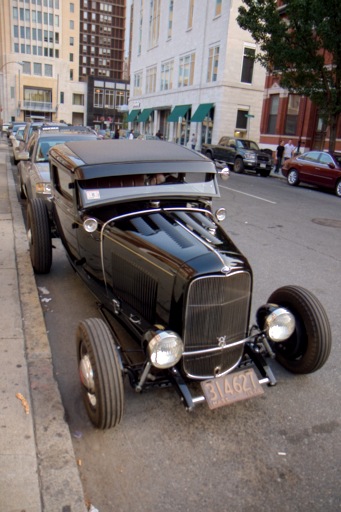}&
    \includegraphics[width=0.29\linewidth]{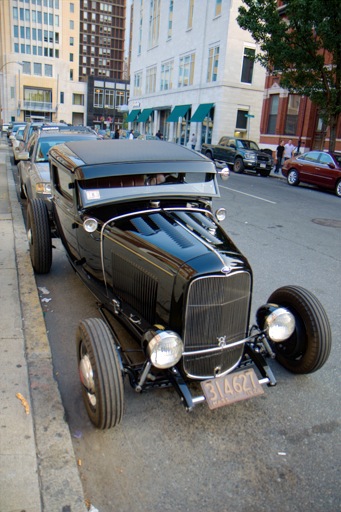}&
    \includegraphics[width=0.29\linewidth]{VisualComparisonMIT5K_paired/a0688_target.jpg}\\
    \includegraphics[width=0.29\linewidth]{VisualComparisonMIT5K_paired/a4994_source.jpg}&
    \includegraphics[width=0.29\linewidth]{VisualComparisonMIT5K_paired/a4994_DPE.jpg}&
    \includegraphics[width=0.29\linewidth]{VisualComparisonMIT5K_paired/a4994_enhanced.jpg}&
    \includegraphics[width=0.29\linewidth]{VisualComparisonMIT5K_paired/a4994_target.jpg}\\
    \includegraphics[width=0.29\linewidth]{VisualComparisonMIT5K_paired/a2846_source.jpg}&
    \includegraphics[width=0.29\linewidth]{VisualComparisonMIT5K_paired/a2846_DPE.jpg}&
    \includegraphics[width=0.29\linewidth]{VisualComparisonMIT5K_paired/a2846_enhanced.jpg}&
    \includegraphics[width=0.29\linewidth]{VisualComparisonMIT5K_paired/a2846_target.jpg}\\
    \includegraphics[width=0.29\linewidth]{VisualComparisonMIT5K_paired/a3245_source.jpg}&
    \includegraphics[width=0.29\linewidth]{VisualComparisonMIT5K_paired/a3245_DPE.jpg}&
    \includegraphics[width=0.29\linewidth]{VisualComparisonMIT5K_paired/a3245_enhanced.jpg}&
    \includegraphics[width=0.29\linewidth]{VisualComparisonMIT5K_paired/a3245_target.jpg}\\
    input & DPE SL & Ours SL & label\\
\end{tabular}
}
\vspace{-0.2cm}
\caption{Visual comparison of learned supervised mappings of our GSGN network against the DPE method on images of the test set of the MIT5K dataset (expert C).}
\label{fig:supplement_visual_comparison_supervised2}
\end{figure*}

Figures~\ref{fig:supplement_visual_comparison_supervised} and ~\ref{fig:supplement_visual_comparison_supervised2} show a visual comparison of learned supervised mappings by our GSGN model against the state-of-the-art method DPE~\cite{DPE} on additional images from the test split of the MIT5K~\cite{fivek} dataset. While the DPE model is able to approximate the target in many cases, our model yields more consistent results and is able to handle a larger variety of input images.

\clearpage

\subsection{Supervised learning - Visual results of multitask enhancement experiment}
\begin{figure*}[hbt!]
\centering
\setlength{\tabcolsep}{1pt}
\resizebox{\linewidth}{!}
{
\begin{tabular}{cccccc}
    \includegraphics[width=0.19\linewidth]{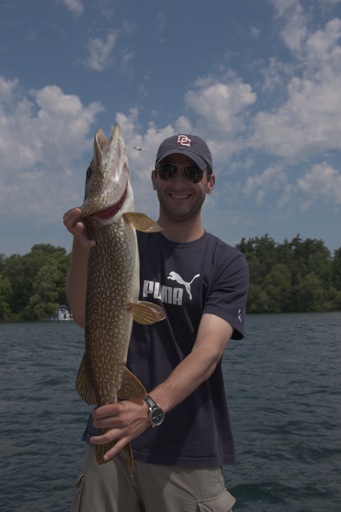}&
    \includegraphics[width=0.19\linewidth]{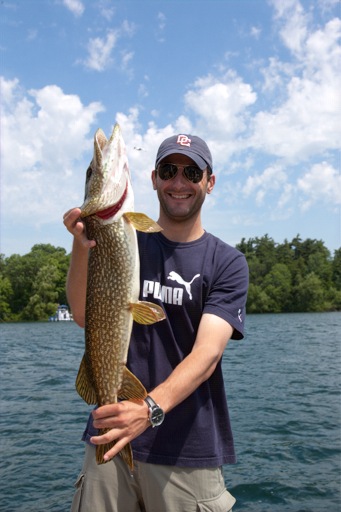}&
    \includegraphics[width=0.19\linewidth]{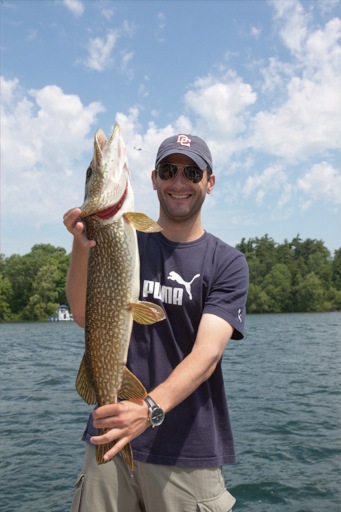}&
    \includegraphics[width=0.19\linewidth]{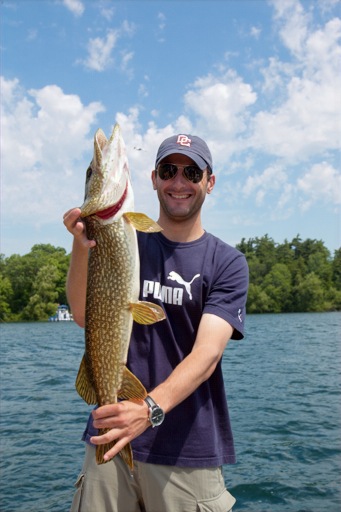}&
    \includegraphics[width=0.19\linewidth]{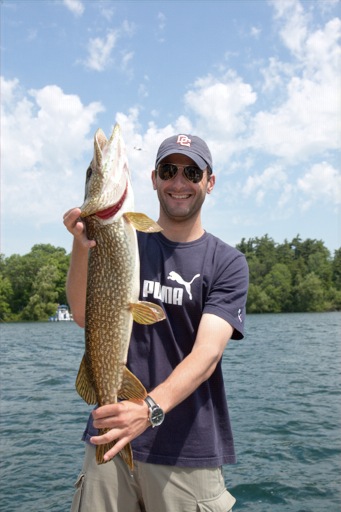}&
    \includegraphics[width=0.19\linewidth]{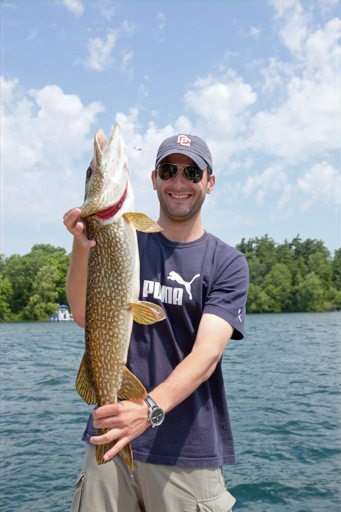}\\
    input & learned A & learned B & learned C & learned D & learned E\\
    &
    \includegraphics[width=0.19\linewidth]{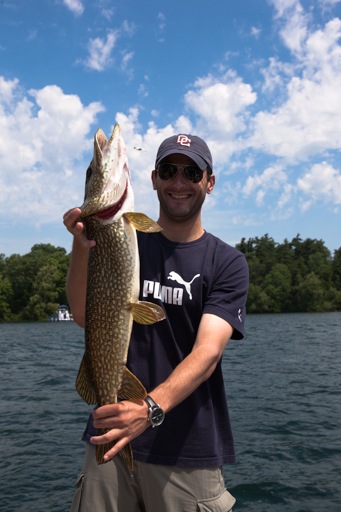}&
    \includegraphics[width=0.19\linewidth]{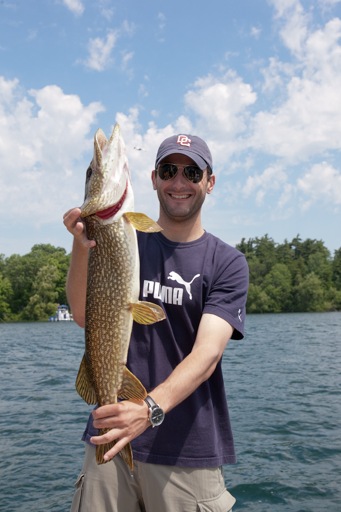}&
    \includegraphics[width=0.19\linewidth]{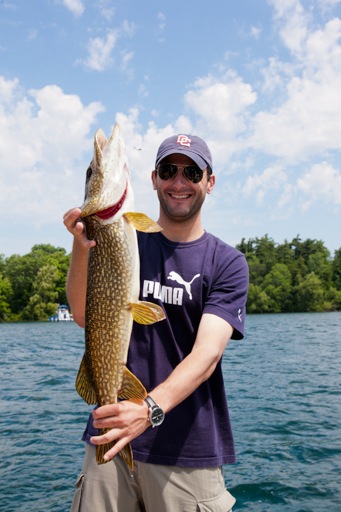}&
    \includegraphics[width=0.19\linewidth]{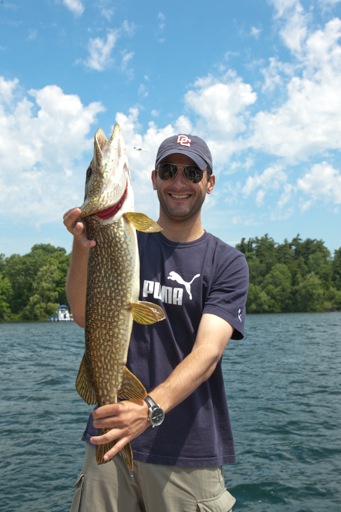}&
    \includegraphics[width=0.19\linewidth]{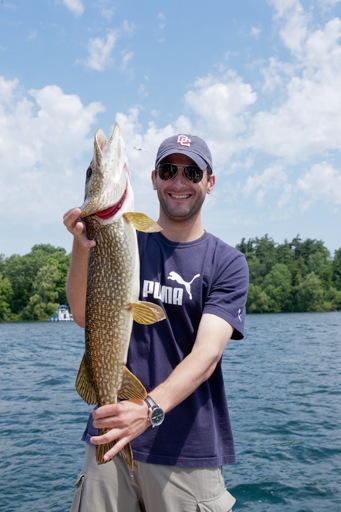}\\
    & expert A & expert B & expert C & expert D & expert E\\
\end{tabular}
}
\vspace{1mm}
\caption{Visual comparison of supervised learned mappings by our MT-GSGN model and expert labels. }
\label{fig:sup_MTGSGN_mit5k_supervised1}
\end{figure*}

\begin{figure*}[hbt!]
\centering
\setlength{\tabcolsep}{1pt}
\resizebox{\linewidth}{!}
{
\begin{tabular}{cccccc}
    \includegraphics[width=0.19\linewidth]{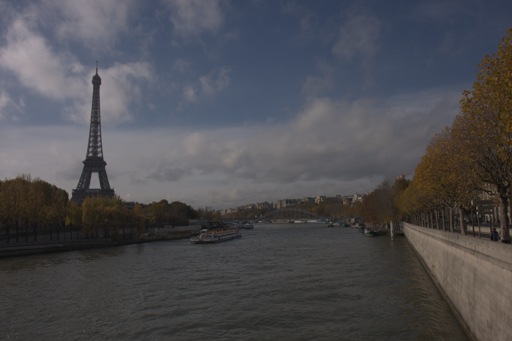}&
    \includegraphics[width=0.19\linewidth]{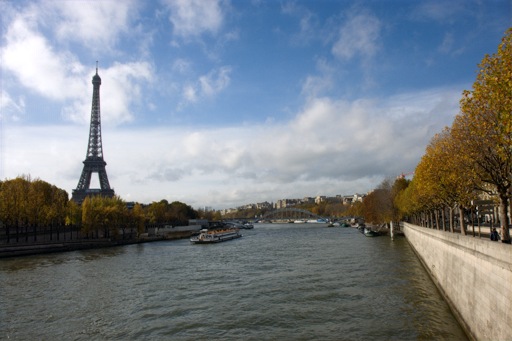}&
    \includegraphics[width=0.19\linewidth]{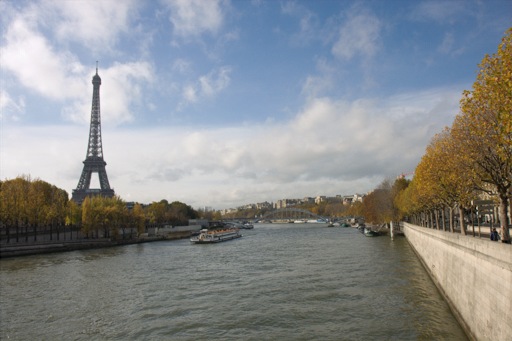}&
    \includegraphics[width=0.19\linewidth]{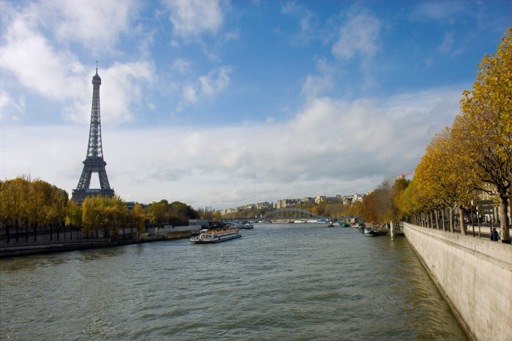}&
    \includegraphics[width=0.19\linewidth]{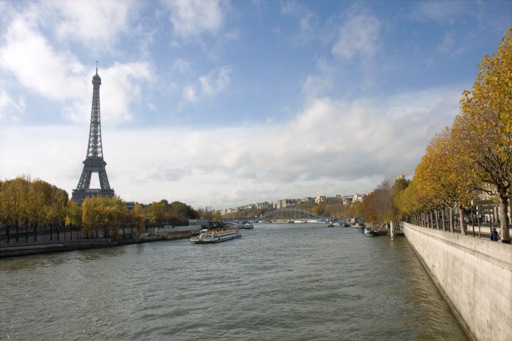}&
    \includegraphics[width=0.19\linewidth]{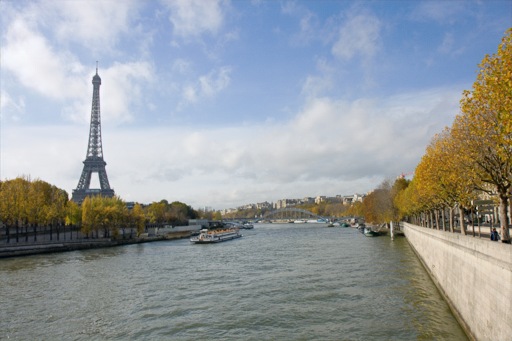}\\
    input & learned A & learned B & learned C & learned D & learned E\\
    &
    \includegraphics[width=0.19\linewidth]{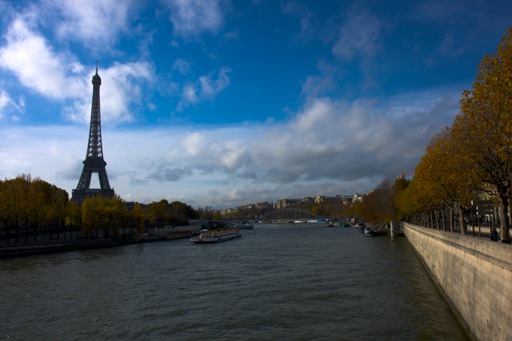}&
    \includegraphics[width=0.19\linewidth]{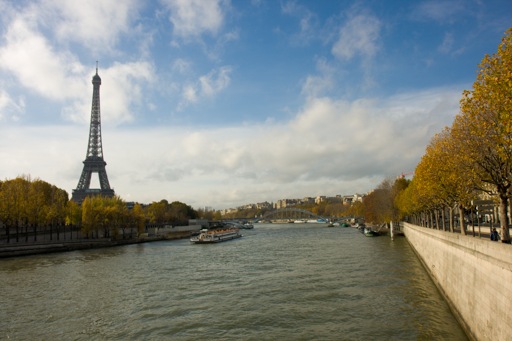}&
    \includegraphics[width=0.19\linewidth]{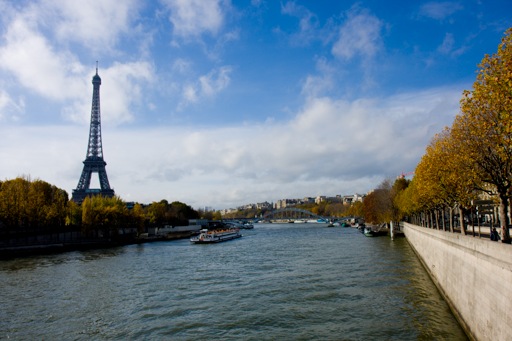}&
    \includegraphics[width=0.19\linewidth]{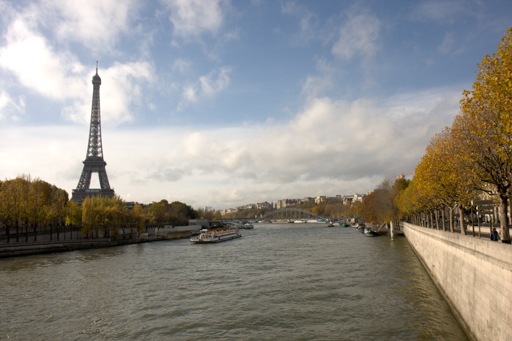}&
    \includegraphics[width=0.19\linewidth]{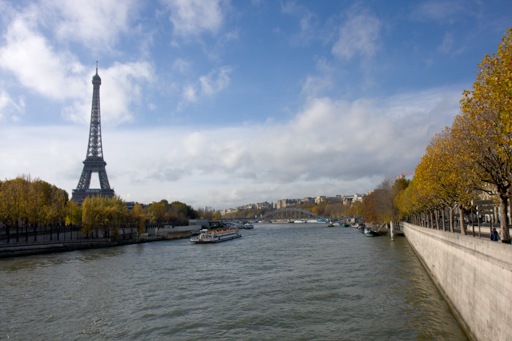}\\
    & expert A & expert B & expert C & expert D & expert E\\
\end{tabular}
}
\vspace{1mm}
\caption{Visual comparison of supervised learned mappings by our MT-GSGN model and expert labels. }
\label{fig:sup_MTGSGN_mit5k_supervised2}
\end{figure*}

Figure~\ref{fig:sup_MTGSGN_mit5k_supervised1} and figure~\ref{fig:sup_MTGSGN_mit5k_supervised2} show results of our MT-GSGN model on the task of supervised multitask learning on the five experts of the MIT5K~\cite{fivek} dataset. The metrics in table 2 in the main paper
show impressive results of the models ability to learn the distribution of the five experts. However, looking at single samples shows little variation between experts in both training data and predictions made by the model. 
Considering that the human retouches of the different experts have limited consistency, this might also mean that a significant portion of an experts style is random.
However, our result still has relevance. Even though the variance between experts on single images is limited, computing the average distance over the whole test set (500 images per expert) in terms of PSNR, SSIM and LPIPS~\cite{LPIPS2018} (table 2 in the paper) shows consistent results and proves that non random aspects of the different expert distributions can be learned by our model.

\FloatBarrier

\subsection{Unsupervised learning - Visual comparison against the State-of-the-art}
\begin{figure*}[ht!]
\centering
\setlength{\tabcolsep}{1pt}
\resizebox{\linewidth}{!}
{
\begin{tabular}{cccc}
    \includegraphics[width=0.29\linewidth]{VisualComparisonMIT5K_paired/a4994_source.jpg}&
    \includegraphics[width=0.29\linewidth]{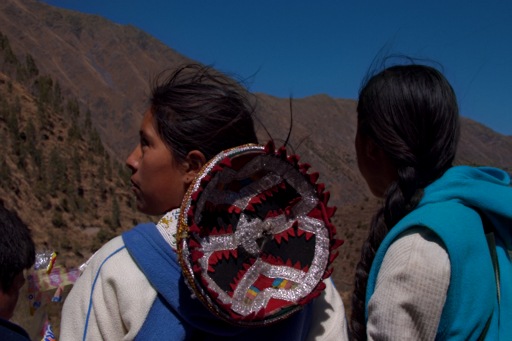}&
    \includegraphics[width=0.29\linewidth]{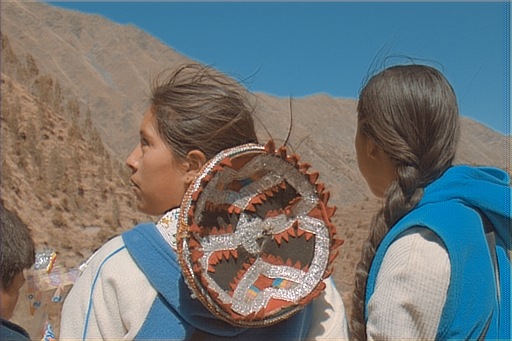}&
    \includegraphics[width=0.29\linewidth]{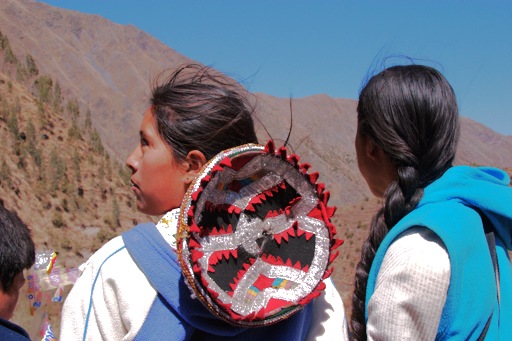}\\
    input & DPE UL & Ours UL & White-Box\\
    \\
    &
    \includegraphics[width=0.29\linewidth]{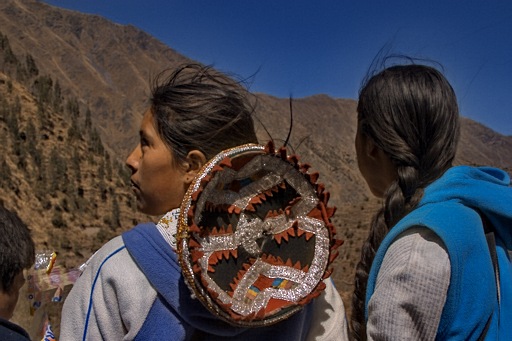}&
    \includegraphics[width=0.29\linewidth]{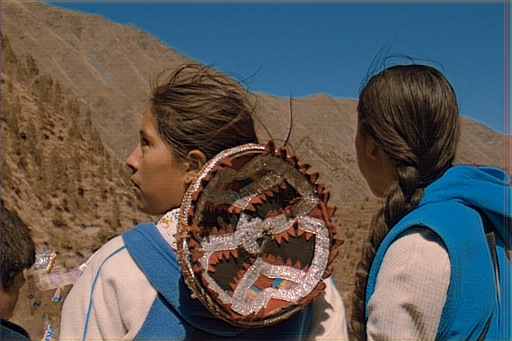}&
    \includegraphics[width=0.29\linewidth]{VisualComparisonMIT5K_paired/a4994_target.jpg}\\
        & DPE HDR & Ours HDR& label\\
\end{tabular}
}
\vspace{1mm}
\caption{Visual comparison of learned unsupervised mappings by our MT-GSGN model trained on the small Flickr Multi-style dataset, against DPE~\cite{DPE} and WhiteBox~\cite{Wbox}.}
\label{fig:sup_MTGSGN_unsup_flickr2}
\end{figure*}

\begin{figure*}[ht!]
\centering
\setlength{\tabcolsep}{1pt}
\resizebox{\linewidth}{!}
{
\begin{tabular}{cccc}
    \includegraphics[width=0.29\linewidth]{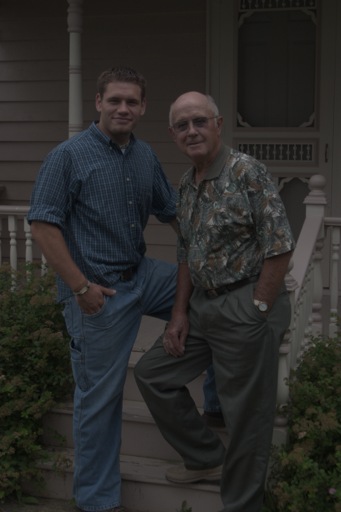}&
    \includegraphics[width=0.29\linewidth]{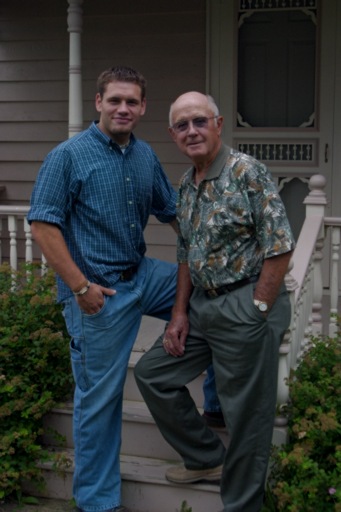}&
    \includegraphics[width=0.29\linewidth]{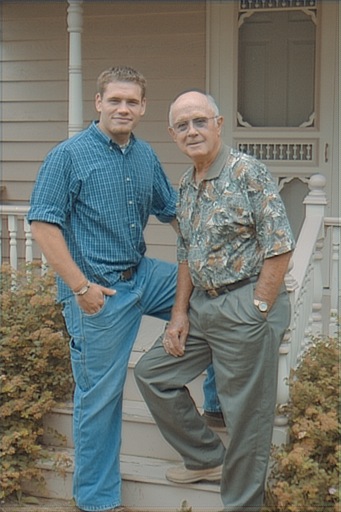}&
    \includegraphics[width=0.29\linewidth]{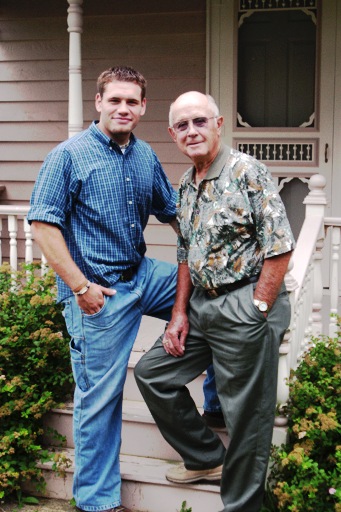}\\
    input & DPE UL & Ours UL & White-Box\\
    \\
    &
    \includegraphics[width=0.29\linewidth]{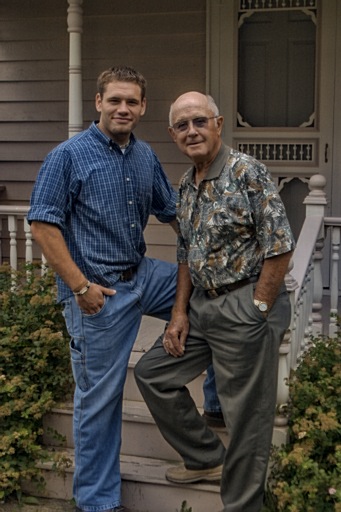}&
    \includegraphics[width=0.29\linewidth]{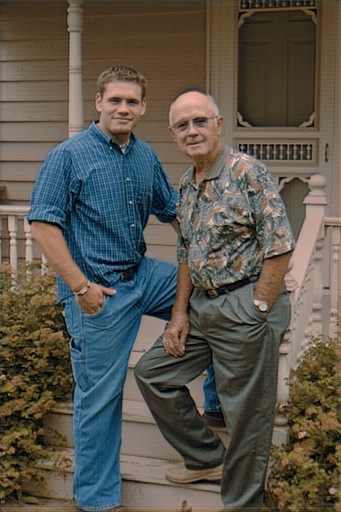}&
    \includegraphics[width=0.29\linewidth]{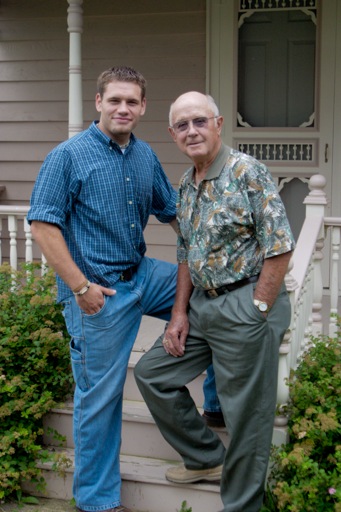}\\
        & DPE HDR & Ours HDR& label\\
\end{tabular}
}
\vspace{1mm}
\caption{Visual comparison of learned unsupervised mappings by our MT-GSGN model trained on the small Flickr Multi-style dataset, against DPE and WhiteBox.}
\label{fig:sup_MTGSGN_unsup_flickr3}
\end{figure*}

\begin{figure*}[ht!]
\centering
\setlength{\tabcolsep}{1pt}
\resizebox{\linewidth}{!}
{
\begin{tabular}{cccc}
    \includegraphics[width=0.29\linewidth]{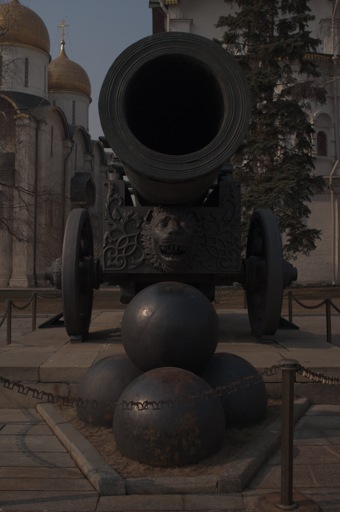}&
    \includegraphics[width=0.29\linewidth]{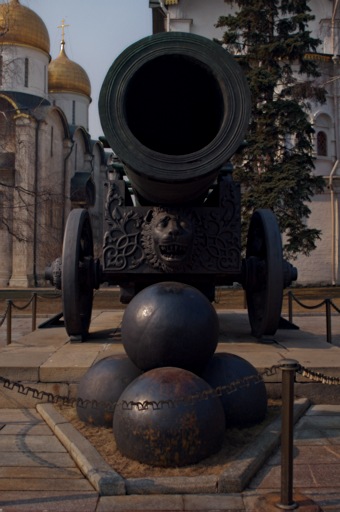}&
    \includegraphics[width=0.29\linewidth]{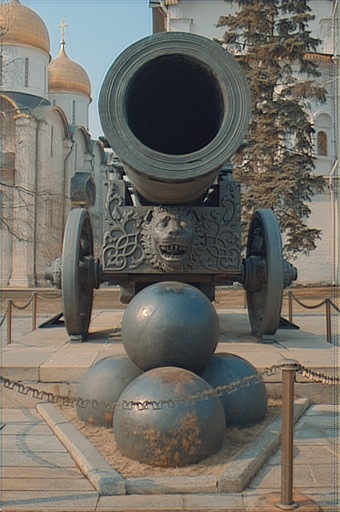}&
    \includegraphics[width=0.29\linewidth]{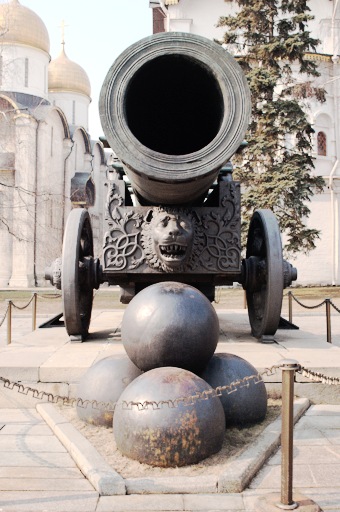}\\
    input & DPE UL & Ours UL & White-Box\\
    \\
    &
    \includegraphics[width=0.29\linewidth]{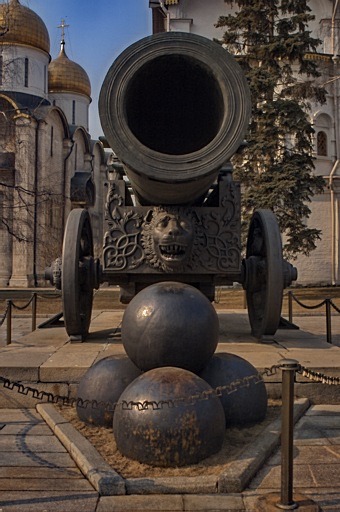}&
    \includegraphics[width=0.29\linewidth]{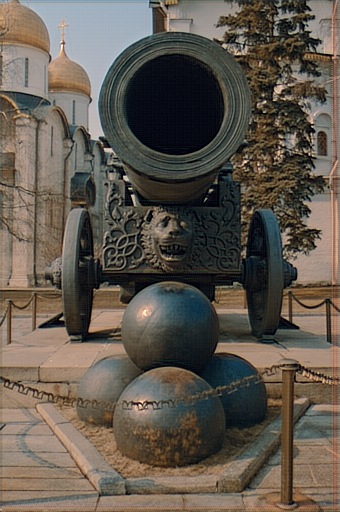}&
    \includegraphics[width=0.29\linewidth]{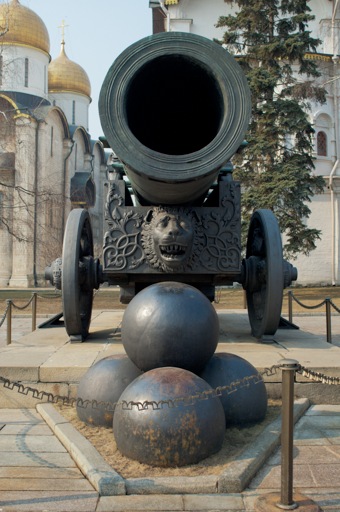}\\
        & DPE HDR & Ours HDR& label\\
\end{tabular}
}
\vspace{1mm}
\caption{Visual comparison of learned unsupervised mappings by our MT-GSGN model trained on the small Flickr Multi-style dataset, against DPE and WhiteBox.}
\label{fig:sup_MTGSGN_unsup_flickr4}
\end{figure*}

\begin{figure*}[ht!]
\centering
\setlength{\tabcolsep}{1pt}
\resizebox{\linewidth}{!}
{
\begin{tabular}{cccc}
    \includegraphics[width=0.29\linewidth]{VisualComparisonMIT5K_paired/a1544_source.jpg}&
    \includegraphics[width=0.29\linewidth]{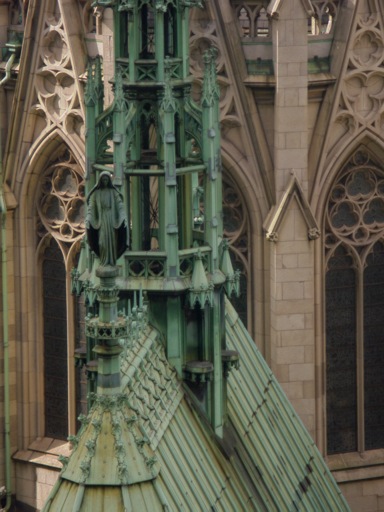}&
    \includegraphics[width=0.29\linewidth]{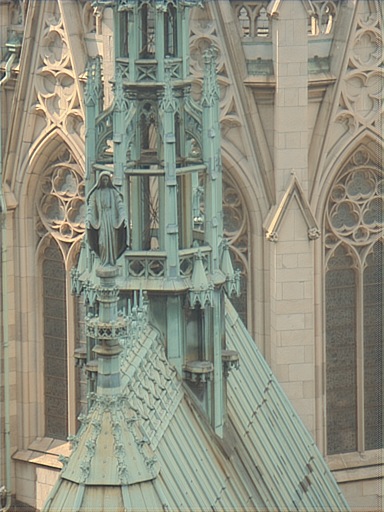}&
    \includegraphics[width=0.29\linewidth]{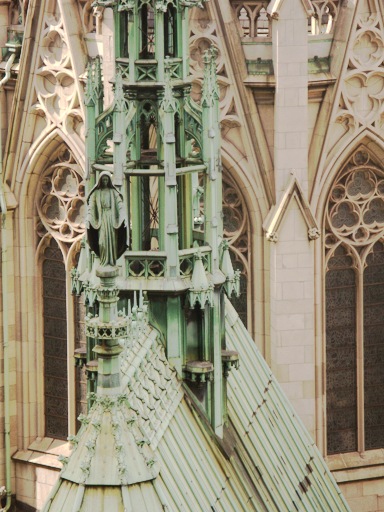}\\
    input & DPE UL & Ours UL & White-Box\\
    \\
    &
    \includegraphics[width=0.29\linewidth]{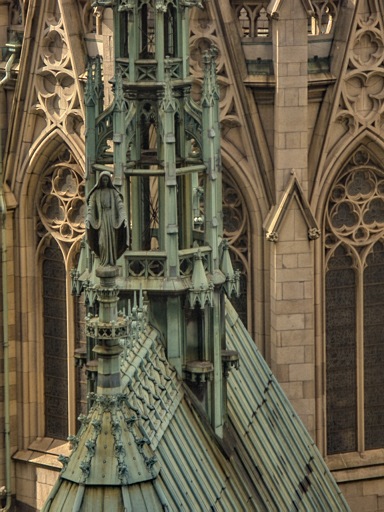}&
    \includegraphics[width=0.29\linewidth]{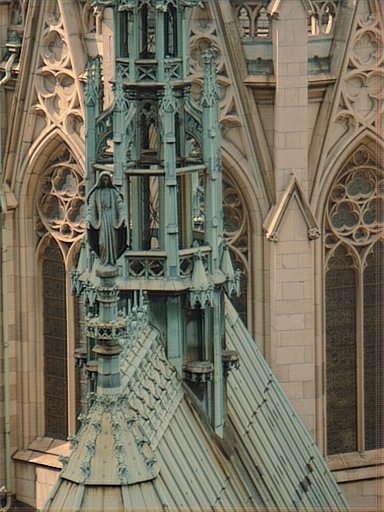}&
    \includegraphics[width=0.29\linewidth]{VisualComparisonMIT5K_paired/a1544_target.jpg}\\
        & DPE HDR & Ours HDR& label\\
\end{tabular}
}
\vspace{1mm}
\caption{Visual comparison of learned unsupervised mappings by our MT-GSGN model trained on the small Flickr Multi-style dataset, against DPE and WhiteBox.}
\label{fig:sup_MTGSGN_unsup_flickr5}
\end{figure*}

\begin{figure*}[ht!]
\centering
\setlength{\tabcolsep}{1pt}
\resizebox{\linewidth}{!}
{
\begin{tabular}{cccc}
    \includegraphics[width=0.29\linewidth]{VisualComparisonMIT5K_paired/a1033_source.jpg}&
    \includegraphics[width=0.29\linewidth]{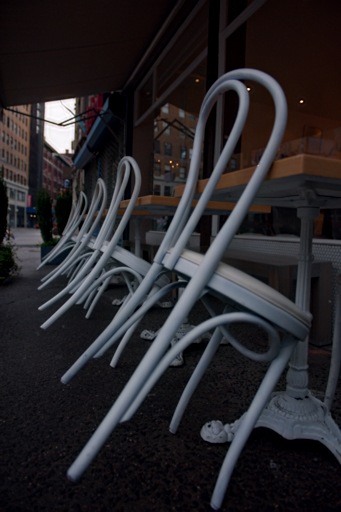}&
    \includegraphics[width=0.29\linewidth]{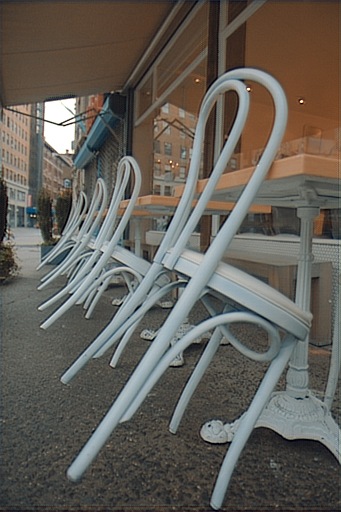}&
    \includegraphics[width=0.29\linewidth]{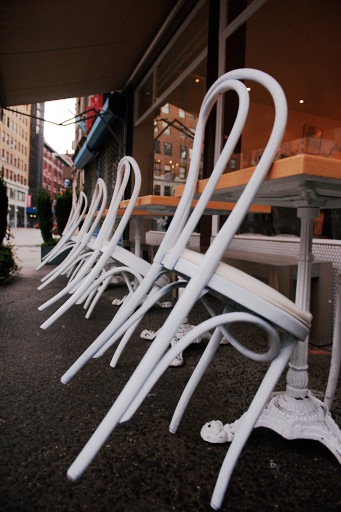}\\
    input & DPE UL & Ours UL & White-Box\\
    \\
    &
    \includegraphics[width=0.29\linewidth]{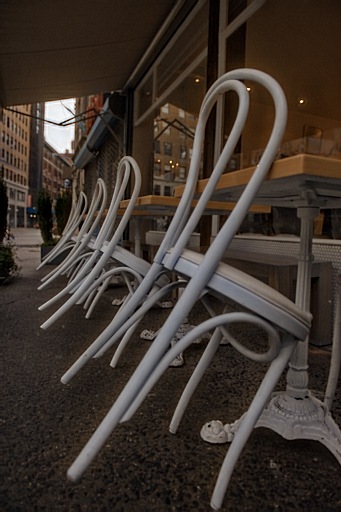}&
    \includegraphics[width=0.29\linewidth]{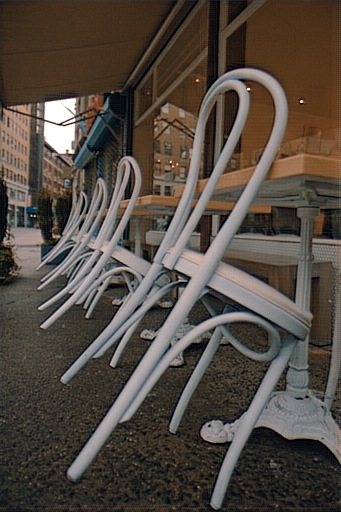}&
    \includegraphics[width=0.29\linewidth]{VisualComparisonMIT5K_paired/a1033_target.jpg}\\
        & DPE HDR & Ours HDR& label\\
\end{tabular}
}
\vspace{1mm}
\caption{Visual comparison of learned unsupervised mappings by our MT-GSGN model trained on the small Flickr Multi-style dataset, against DPE and WhiteBox.}
\label{fig:sup_MTGSGN_unsup_flickr6}
\end{figure*}

\begin{figure*}[ht!]
\centering
\setlength{\tabcolsep}{1pt}
\resizebox{\linewidth}{!}
{
\begin{tabular}{cccc}
    \includegraphics[width=0.29\linewidth]{VisualComparisonMIT5K_paired/a0503_source.jpg}&
    \includegraphics[width=0.29\linewidth]{VisualComparisonFlickr/a0503_DPE_mit.jpg}&
    \includegraphics[width=0.29\linewidth]{VisualComparisonFlickr/a0503_L010_enh.jpg}&
    \includegraphics[width=0.29\linewidth]{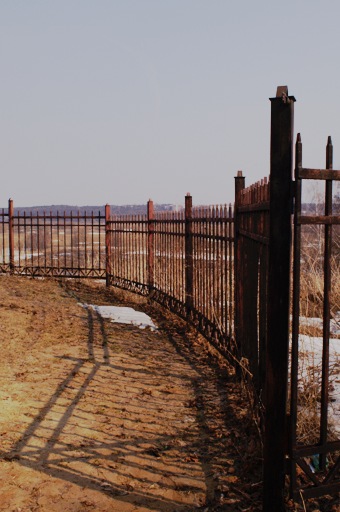}\\
    input & DPE UL & Ours UL & White-Box\\
    \\
    &
    \includegraphics[width=0.29\linewidth]{VisualComparisonFlickr/a0503_DPE_HDR.jpg}&
    \includegraphics[width=0.29\linewidth]{VisualComparisonFlickr/a0503_L100_enh.jpg}&
    \includegraphics[width=0.29\linewidth]{VisualComparisonMIT5K_paired/a0503_target.jpg}\\
        & DPE HDR & Ours HDR& label\\
\end{tabular}
}
\vspace{1mm}
\caption{Visual comparison of learned unsupervised mappings by our MT-GSGN model trained on the small Flickr Multi-style dataset, against DPE and WhiteBox.}
\label{fig:sup_MTGSGN_unsup_flickr7}
\end{figure*}

\begin{figure*}[ht!]
\centering
\setlength{\tabcolsep}{1pt}
\resizebox{\linewidth}{!}
{
\begin{tabular}{cccc}
    \includegraphics[width=0.29\linewidth]{VisualComparisonMIT5K_paired/a0688_source.jpg}&
    \includegraphics[width=0.29\linewidth]{VisualComparisonFlickr/a0688_DPE_mit.jpg}&
    \includegraphics[width=0.29\linewidth]{VisualComparisonFlickr/a0688_L010_enh.jpg}&
    \includegraphics[width=0.29\linewidth]{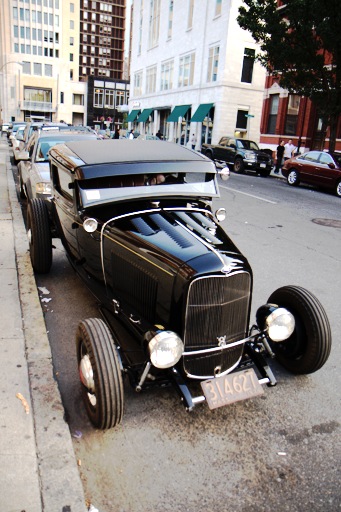}\\
    input & DPE UL & Ours UL & White-Box\\
    \\
    &
    \includegraphics[width=0.29\linewidth]{VisualComparisonFlickr/a0688_DPE_HDR.jpg}&
    \includegraphics[width=0.29\linewidth]{VisualComparisonFlickr/a0688_L100_enh.jpg}&
    \includegraphics[width=0.29\linewidth]{VisualComparisonMIT5K_paired/a0688_target.jpg}\\
        & DPE HDR & Ours HDR& label\\
\end{tabular}
}
\vspace{1mm}
\caption{Visual comparison of learned unsupervised mappings by our MT-GSGN model trained on the small Flickr Multi-style dataset, against DPE and WhiteBox.}
\label{fig:sup_MTGSGN_unsup_flickr8}
\end{figure*}

\begin{figure*}[ht!]
\centering
\setlength{\tabcolsep}{1pt}
\resizebox{\linewidth}{!}
{
\begin{tabular}{cccc}
    \includegraphics[width=0.29\linewidth]{VisualComparisonMIT5K_paired/a1581_source.jpg}&
    \includegraphics[width=0.29\linewidth]{VisualComparisonFlickr/a1581_DPE_mit.jpg}&
    \includegraphics[width=0.29\linewidth]{VisualComparisonFlickr/a1581_L010_enh.jpg}&
    \includegraphics[width=0.29\linewidth]{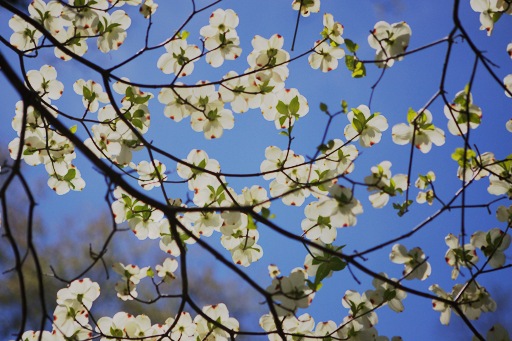}\\
    input & DPE UL & Ours UL & White-Box\\
    \\
    &
    \includegraphics[width=0.29\linewidth]{VisualComparisonFlickr/a1581_DPE_HDR.jpg}&
    \includegraphics[width=0.29\linewidth]{VisualComparisonFlickr/a1581_L100_enh.jpg}&
    \includegraphics[width=0.29\linewidth]{VisualComparisonMIT5K_paired/a1581_target.jpg}\\
        & DPE HDR & Ours HDR& label\\
\end{tabular}
}
\vspace{1mm}
\caption{Visual comparison of learned unsupervised mappings by our MT-GSGN model trained on the small Flickr Multi-style dataset, against DPE and WhiteBox.}
\label{fig:sup_MTGSGN_unsup_flickr9}
\end{figure*}

As the MIT5K~\cite{fivek} dataset has limited variation between the different experts in terms of style, we decided to train our MT-GSGN model on a custom made small flickr multi style dataset. Figures~\ref{fig:sup_MTGSGN_unsup_flickr2} to~\ref{fig:sup_MTGSGN_unsup_flickr9} show a visual comparison of learned unsupervised mappings by our MT-GSGN model against the state-of-the-art unsupervised methods DPE~\cite{DPE} and WhiteBox~\cite{Wbox}.
WhiteBox yiels good results but often pushes the contrast too far. Our UL style result shows bright images but with limited contrast. The two HDR methods (DPE HDR and Ours HDR) both show visually pleasing results. While DPE has problems with artefacts in high contrast areas, especially on the border between the blue sky and the motiv, our method sometimes shows pixel level artefacts. We expect that these can be improved or resolved by fine tuning the training, as no extensive hyper-parameter search was performed in this example.
Note that the two styles of the DPE method require two separate models and trainings, while our MT-GSGN method learns multiple styles in a single training using a compact model.

\FloatBarrier

\begin{figure*}[ht!]
\centering
\setlength{\tabcolsep}{1pt}
\resizebox{\linewidth}{!}
{
\begin{tabular}{cccc}
    \includegraphics[width=0.29\linewidth]{VisualComparisonMIT5K_paired/a2846_source.jpg}&
    \includegraphics[width=0.29\linewidth]{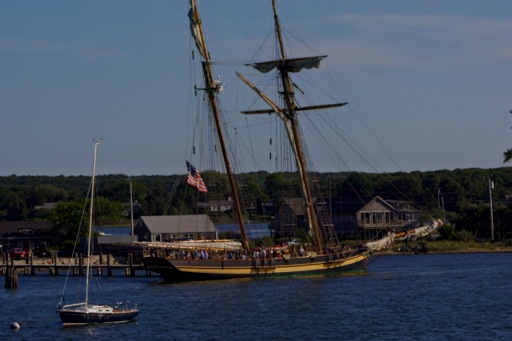}&
    \includegraphics[width=0.29\linewidth]{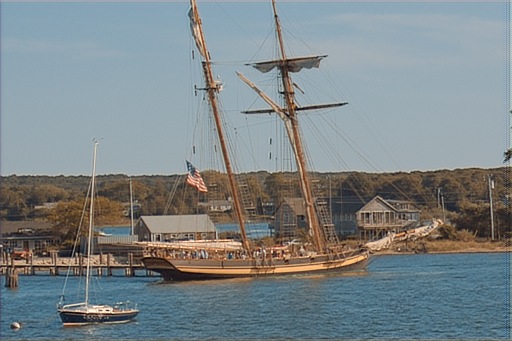}&
    \includegraphics[width=0.29\linewidth]{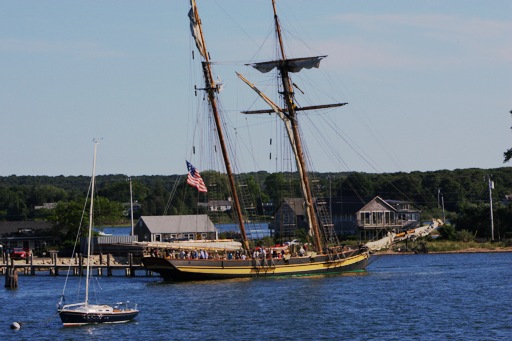}\\
    input & DPE UL & Ours UL & White-Box\\
    \\
    &
    \includegraphics[width=0.29\linewidth]{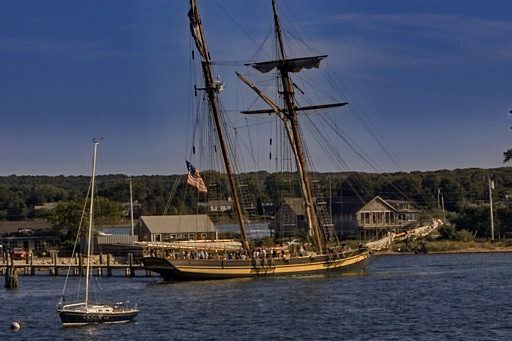}&
    \includegraphics[width=0.29\linewidth]{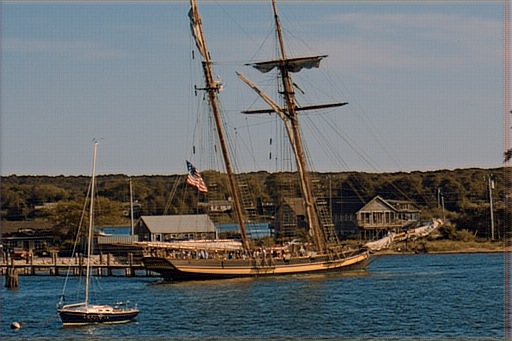}&
    \includegraphics[width=0.29\linewidth]{VisualComparisonMIT5K_paired/a2846_target.jpg}\\
        & DPE HDR & Ours HDR& label\\
\end{tabular}
}
\vspace{1mm}
\caption{Visual comparison of learned unsupervised mappings by our MT-GSGN model trained on the small Flickr Multi-style dataset, against DPE and WhiteBox.}
\end{figure*}

\end{document}